\pdfoutput=1 
\documentclass[lettersize,journal]{IEEEtran}
\usepackage{amsmath,amsfonts}
\usepackage{algpseudocode}
\usepackage{algorithm}
\usepackage{array}
\usepackage[caption=false,font=normalsize,labelfont=sf,textfont=sf]{subfig}
\usepackage{textcomp}
\usepackage{stfloats}
\usepackage{url}
\usepackage{verbatim}
\usepackage{cite}
\usepackage{euscript}
\usepackage{gensymb}
\usepackage{graphicx}
\usepackage{pgfplots}
\usepackage{xcolor}
\usepackage{hyperref}
\usepackage{academicons}
\usepackage{svg}

\newcommand{\orcid}[1]{\href{https://orcid.org/#1}{\includegraphics[width=10pt]{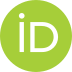}}}

\DeclareMathOperator*{\argmin}{arg\,min}

\pgfplotsset{width=8cm,compat=1.9}

\begin{document}

\title{PU-Ray: Domain-Independent Point Cloud Upsampling via  Ray Marching on Neural Implicit Surface}

\author{
    \thanks{Manuscript received ...}

    Sangwon Lim \orcid{0009-0003-8270-2267},
    \thanks{Sangwon Lim is with the Centre for Smart Transportation (CST) and the Department of Computing Science at the University of Alberta, Edmonton, AB, Canada (e-mail: sangwon3@ualberta.ca).}
  Karim El-Basyouny,
    \thanks{Karim El-Basyouny is with the Centre for Smart Transportation (CST) and the Department of Civil and Environmental Engineering, University of Alberta, Edmonton, AB, Canada T6G 1H9 (e-mail: basyouny@ualberta.ca).}
  and
  Yee Hong Yang \orcid{0000-0002-7194-3327}
    \thanks{Yee Hong Yang is with the Department of Computing Science, University of Alberta, Edmonton, AB, Canada T6G 2E8 (e-mail: herberty@ualberta.ca).}
}



\renewcommand\fbox{\fcolorbox{red}{white}}
\setlength{\fboxrule}{2pt} 

\newcommand\submittedtext{%
  \footnotesize This work has been submitted to the IEEE for possible publication. Copyright may be transferred without notice, after which this version may no longer be accessible.}

\newcommand\submittednotice{%
\begin{tikzpicture}[remember picture,overlay]
\node[anchor=south,yshift=10pt] at (current page.south) {\fbox{\parbox{\dimexpr0.65\textwidth-\fboxsep-\fboxrule\relax}{\submittedtext}}};
\end{tikzpicture}%
}

\maketitle
\submittednotice

\begin{abstract}
While recent advancements in deep-learning point cloud upsampling methods have improved the input to intelligent transportation systems, they still suffer from issues of domain dependency between synthetic and real-scanned point clouds. This paper addresses the above issues by proposing a new ray-based upsampling approach with an arbitrary rate, where a depth prediction is made for each query ray and its corresponding patch. Our novel method simulates the sphere-tracing ray marching algorithm on the neural implicit surface defined with an unsigned distance function (UDF) to achieve more precise and stable ray-depth predictions by training a point-transformer-based network. The rule-based mid-point query sampling method generates more evenly distributed points without requiring an end-to-end model trained using a nearest-neighbor-based reconstruction loss function, which may be biased towards the training dataset. Self-supervised learning becomes possible with accurate ground truths within the input point cloud. The results demonstrate the method's versatility across domains and training scenarios with limited computational resources and training data. Comprehensive analyses of synthetic and real-scanned applications provide empirical evidence for the significance of the upsampling task across the computer vision and graphics domains to real-world applications of ITS.
\end{abstract}


\section{Introduction}

\IEEEPARstart{L}{iDAR} point clouds provide an understanding of the surrounding environment's surfaces and have been used in applications in the intelligent transportation systems (ITS) domain, ranging from tasks directly related to vehicle controls, such as shape classification, object detection, and point cloud segmentation \cite{point_cloud_survey}, to 3D reconstruction related tasks of mapping and surveying \cite{lidar_performance, completion}. Point cloud upsampling is a task in 3D reconstruction for ITS that can lower the cost of memory storage and sensor requirements. For example, upsampling the data from an HDL-64E sensor \cite{velodyne} can potentially achieve a similar or better quality than that of the state-of-the-art (SOTA) Velodyne Alpha Prime VLS-128 sensor within the maximum operation range, which shows significant performance improvements in autonomous driving \cite{lidar_performance}.

A few studies have addressed the upsampling problem in the domain of ITS \cite{rbf, fusion_review, pseudo-lidar, event_enhance}. However, such enhancement solutions are restricted to specific problems or rely on additional sensors. Studies in range image super-resolution \cite{super_resolution, pseudo-lidar, kwon2022implicit, sgsr}, LiDAR upsampling \cite{lidar_upsampling} and LiDAR completion \cite{ultralidar} do not account for 3D densities. Concurrently, many computer vision studies \cite{pu_net, mpu, pu_gan, pu_gcn, sspu, spu_net, pugac, pu_transformer, gradpu, sapcu} have focused on performance improvements to existing benchmarks. However, the output point distribution is not continuous when the above upsampling methods \cite{pu_net, mpu, pu_gan, pu_gcn, sspu, spu_net, pugac, pu_transformer, gradpu} are applied to real-scanned LiDAR data with many local density mismatches due to the nearest-neighbor-based reconstruction loss functions, such as Chamfer Distance (CD). Also, encoding the entire object shape \cite{pu_net, pu_gan, pu_gcn, sspu, sapcu} may cause domain dependency when encountering an unseen object. Such end-to-end behaviours could pose issues for 3D reconstruction for ITS because real-scanned data include many different objects, and defining the bounding space of the infrastructural environment is difficult. At the same time, the upsampled points generated near the known input points are less important for 3D reconstruction. Additionally, many methods \cite{pu_net, pu_gan, pu_gcn, pu_transformer, spu_net} upsample point clouds with a fixed scaling rate, $r$. Thus, a different model has to be trained to upsample with a different scaling rate. Combined with the issues of end-to-end strategies, the fixed rate causes inflexibility without the freedom of ROI definition and output density. The observations above motivate the proposed method to move away from end-to-end strategies with CD-based loss functions and a fixed upsampling rate.

\IEEEpubidadjcol

\begin{figure}[!t]
\centering
\includegraphics[width=\columnwidth]{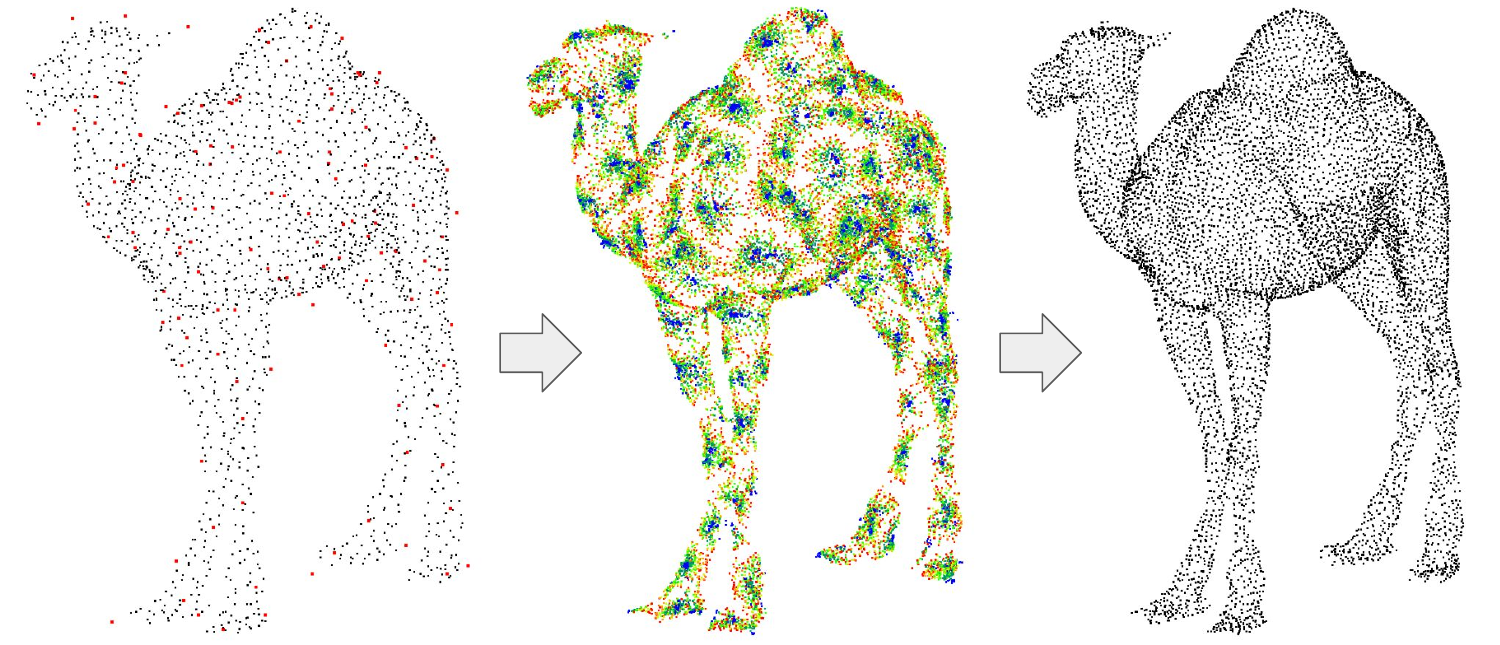}
\caption{Given the sparse input (left), through the ray marching algorithm (middle), the final dense output (right) is achieved. Points on the implicit surface are coloured to represent marching steps from the earlier (blue) to the later (red) steps. The red dots on the left are the initial query ray origins.}
\label{fig:intro}
\end{figure}

\begin{figure*}[!t]
\centering
\includegraphics[width=\textwidth]{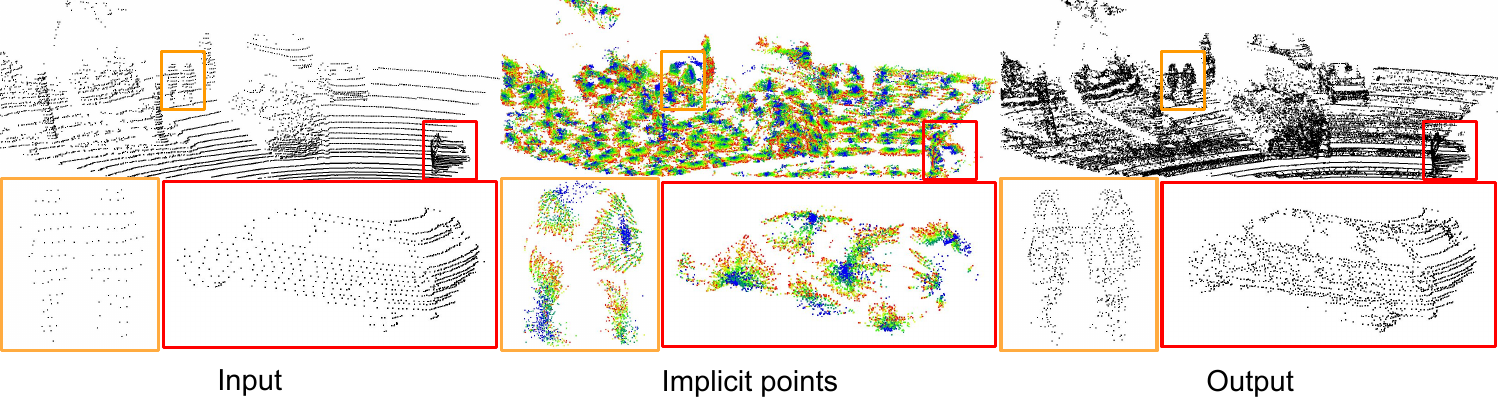}
\caption{Demonstration of PU-Ray through the input, implicit points, and output on a KITTI-360 \cite{kitti360} snippet scene.}
\label{fig:kitti}
\end{figure*}

This paper addresses the listed issues with ray-depth prediction via sphere tracing \cite{sphere_tracing} on the neural implicit surface,  defined using a point-transformer-based \cite{point_transformer} network, for domain-independent point cloud upsampling. Sphere tracing \cite{sphere_tracing} cannot be applied directly to point clouds, particularly with missing information between points. To overcome this constraint, we propose a ray marching module that learns the neural implicit surface of a point cloud patch. Applying the proposed ray-based upsampling method has several advantages over existing end-to-end models. First, the implicit surface of a patch efficiently compresses large 3D data into a few function parameters, and it is domain-independent compared to object-based representations that may be biased toward the dataset's context. Moreover, the upsampling rate is determined by the number of rays generated, enabling upsampling with different rates. The direction of a query ray can be generated to focus on the ROIs. Lastly, unlike existing methods, the training objective and ground truth are clearly stated. Given a patch, the model outputs the depth prediction of a query ray. This allows the retrieval of ground truth in limited situations. The ray can be determined from an existing point's direction and depth from the query ray origin, making self-supervised learning possible. The more straightforward task also reduces the model size. Our contributions are as follows:

\begin{enumerate}
\item{Our method adopts the sphere tracing \cite{sphere_tracing} on the neural} implicit surface,  yielding precise depth predictions while allowing supervised and self-supervised training.
\item{The domain-independent ray-based upsampling with an arbitrary rate is versatile across synthetic to real-scanned and from object-level to infrastructure-level point clouds. Our method shows SOTA performances and model size compression.}
\item{The experimental results on novel datasets consisting of LiDAR point clouds in urban and highway environments provide empirical support for the potential of upsampling real-scanned point clouds.}
\end{enumerate}

\section{Related work}
\subsection{Point Cloud Upsampling}
\subsubsection{Deep-learning Methods}
Before the advent of deep-learning point cloud upsampling methods, edge-aware resampling (EAR) \cite{ear} held SOTA results by projecting points to a latent surface. However, the method requires carefully tuned neighbourhood radius and angle parameters, which vary for different point clouds or regions in a point cloud depending on the point density for accurate surface normal estimation. Thus, its application to multiple point clouds or real-scanned data is difficult, if not impossible, to automate. This inspired the first deep-learning method, PU-Net \cite{pu_net}, adopting the feature encoding method of PointNet++ \cite{pointnet++}. Many earlier point cloud upsampling methods have a fixed upsampling rate and use end-to-end training by object shape encoding, upsampling points with feature expansion and tensor reshaping to fit the ground truth upsampled point clouds \cite{pu_net, mpu, pu_gan, pu_gcn, sspu, spu_net, pugac, pu_transformer}. However, domain-dependent and object-based upsampling with a fixed rate cause inflexibility, which motivates the following studies to address the issues. Unlike 2D image super-resolution, upsampling an unordered and unstructured point cloud does not have a fixed ground truth for the corresponding low-density input. Therefore, many methods rely on nearest-neighbor-based reconstruction loss functions, e.g. Chamfer Distance (CD) and Hausdorff distance (HD) \cite{pu-review, mpu, surface-reconstruction} or their modifications. Combined with end-to-end learning, the upsampling performance relies on the input point distribution. Hence, many upsampling models, which are trained on synthetic datasets, have the domain dependency issue. Wu et al. \cite{DCD} argue the ambiguity of the CD loss function for training a point generation model because it ignores all other points but the nearest point in the target point cloud.

\subsubsection{Neural Implicit Surface}
Sphere tracing \cite{sphere_tracing} on neural implicit surfaces \cite{dist, ndf, csp} have been proposed inspired by DeepSDF \cite{deepsdf} thanks to its capabilities in compressing precise complex shapes. Object-based implicit surfaces may have domain dependency due to shape encoding, and the training is restricted to supervised learning requiring ground truth arbitrary points' distances to the surface. They also use sphere tracing \cite{sphere_tracing} exclusively for 2D rendering. Methods with arbitrary upsampling rates generally have seed points, which are the initial samples before refinement using an implicit surface \cite{sapcu, gradpu}. SAPCU \cite{sapcu} voxelizes the 3D space uniformly and uses the center points of voxels as seeds. A drawback is the high computational cost of generating seed points. Indeed, the farthest point sampling (FPS) \cite{pointnet++} could be applied to improve performance. Grad-PU \cite{gradpu} adopts the midpoint algorithm in local neighbourhoods and achieves SOTA performance.

\subsubsection{Self-supervised Learning}
Several studies have demonstrated the potential of self-supervised learning for point cloud upsampling \cite{sspu, spu_net, sapcu}. The first two methods, SSPU and SPU-Net \cite{sspu, spu_net}, follow a similar point generation method with earlier models using feature expansion and tensor reshaping. On the other hand, SAPCU \cite{sapcu} allows for an arbitrary upsampling rate with seed point generation. Although they have a similar problem statement to our study, the main difference is that they must calculate the pseudo-ground-truth for point projection on the implicit surface. In contrast, our method has accurate ground truths based on existing points in the input point cloud. 

\subsubsection{Applications to LiDAR Scans}
Datasets for point cloud upsampling are less standardized for applications using LiDAR scans than synthetic datasets used in conventional computer vision studies. For example, Li et al. \cite{rbf} propose a problem-specific interpolation-based point cloud density enhancement method for long-range pedestrian detection with a sparse point density \cite{rbf}. Camera-LiDAR fusion methods \cite{fusion_review, pseudo-lidar, event_enhance} are suggested to obtain denser outputs. As an alternative direction, super-resolution on a 2D range image or depth map has been proposed \cite{super_resolution, pseudo-lidar, kwon2022implicit, sgsr}. Although the coarse-to-fine objective of super-resolution is similar to that of point cloud upsampling, the resulting point distribution from super-resolution may not be uniform in the 3D space, as the task focuses on obtaining a high resolution in the 2D space. On the other hand, LiDAR Upsampling by Savkin et al. \cite{lidar_upsampling} and LiDAR Completion by Xiong et al. \cite{ultralidar} have attempted to generate an output point cloud directly from the input point cloud without an intermediate 2D representation. However, the objectives of both methods are similar to super-resolution methods, as the ground truths are high-resolution LiDAR scans due to the difficulty in collecting dense point clouds. Our proposed method solves this issue with ray-based upsampling.

Point cloud completion \cite{completion} is a closely related task of upsampling by completing the surface shape by generating points. The comprehensive review by Fei et al. \cite{completion} on point cloud completion emphasizes the importance of the 3D reconstruction task in the domain of ITS by lowering storage cost and sensor requirements reduction. Most existing methods are object-based completion methods and cannot be applied to real scenes. Similar emphases are found in other previous real scene 3D reconstruction studies for ITS \cite{rbf, fusion_review, pseudo-lidar, event_enhance, super_resolution, kwon2022implicit, sgsr, lidar_upsampling, ultralidar}.

\subsection{Attention and transformer modules for point clouds}
The transformer architecture has been extensively applied to other fields since the first method was proposed, showing impressive performance in natural language processing \cite{vaswani2017attention}. In particular, the transformer has been employed in unorganized point cloud data with great potential, as shown in recent work \cite{pct, point_transformer, lft}. The main idea involves calculating point attention within a small patch of neighbouring points to extract the relationships between sampled points. The practice is advantageous because the 3D coordinates of points naturally convey positional information without needing the positional encoding step. Attention calculation has two major variants. Guo et al. \cite{pct} employ the traditional transformer's scalar attention using dot products. In contrast, Zhao et al. \cite{point_transformer} utilize the vector representation of attention through element-wise subtraction. Applying self-attention was first introduced to point cloud upsampling by Li et al. \cite{pu_gan}. One of the latest advancements in point cloud upsampling, with SOTA performance, also relies on the transformer architecture \cite{pu_transformer}. 

\section{Methodology}
Our proposed method, PU-Ray, performs the sphere tracing algorithm \cite{sphere_tracing} on the neural implicit surface defined with unsigned distance function (UDF). With our novel loss functions, PU-Ray effectively trains a precise depth prediction model through sphere-tracing \cite{sphere_tracing} while using a small number of parameters. The trained UDF is combined with our query ray generation algorithm for upsampling. In the following subsections, the problem statement explicitly explains how the implicit surface is defined. The query ray generation algorithms and the network architecture are discussed.

\subsection{Problem statement}
\label{sect:problem_statement}

\begin{figure}[!htb]
\centering
\includegraphics[width=\columnwidth]{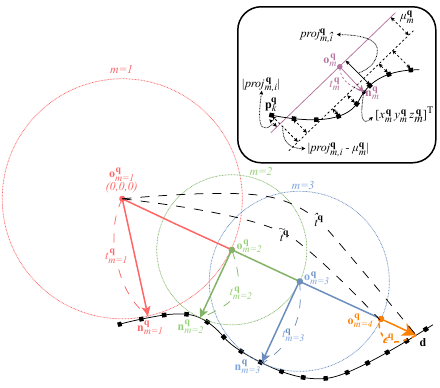}
\caption{Visual demonstration of the ray marching method on the implicit surface. The origin, $\textbf{o}^{\textbf{q}}_m$, is updated at every marching step, $m$, given the nearest distance to the implicit surface, $t^{\textbf{q}}_m$. The final query ray depth, $\hat{t}^{\textbf{q}}$, is the sum of the cumulative depth, $\Tilde{t}^{\textbf{q}}$, and an offset, $\epsilon^{\textbf{q}}$. The inset at the top-right is a visual aid to the nearest point search of $(x^{\textbf{q}}_m, y^{\textbf{q}}_m, z^{\textbf{q}}_m)$ on the implicit surface. The purple plane is defined with $\textbf{o}^{\textbf{q}}_m$ and its implicit nearest direction, $\textbf{n}^{\textbf{q}}_m$. The projection distances to the approximated tangent plane (dotted straight line) are defined by $|proj^{\textbf{q}}_{m, i} - \mu^{\textbf{q}}_m|$'s. The nearest implicit distance, $t^{\textbf{q}}_m$, is approximated by the projection distance of the nearest point in the patch defined by $proj^{\textbf{q}}_{m, \hat{i}}$.}
\label{fig:ray_marching}
\end{figure}

\begin{algorithm}
\caption{Ray marching on point cloud}
\label{alg:ray_marching}

\begin{algorithmic}  
\Require Query ray direction,  $\textbf{d}^{\textbf{q}}$ \Comment{Sect. \ref{sect:problem_statement} \& \ref{sect:query-patch}}
\Require Patch, $P^{\textbf{q}}$ \Comment{Sect. \ref{sect:problem_statement} \& \ref{sect:query-patch}}
\Require $MLP_F(\cdot)$, $MLP_I(\cdot)$ and $MLP_\epsilon^{\textbf{q}}(\cdot)$ \Comment{Sect. \ref{sect:network}}
\Require Point Transformer Module, $PT(\cdot)$ \Comment{Sect. \ref{sect:feature}}
\Require Cross Attention Module, $CA(\cdot)$ \Comment{Sect. \ref{sect:feature}}
\Ensure Maximum number of marching steps, $M\geq 0$
\State $m \leftarrow 1$ \Comment{Initialize marching step index}
\State $\textbf{o}^{\textbf{q}}_m \leftarrow [0 \: 0 \: 0]^T$ \Comment{Initialize origin}
\State $\Tilde{t}^{\textbf{q}} \leftarrow 0$ \Comment{Initialize cumulative depth}

\State $F^{\textbf{q}}_P \leftarrow PT(MLP_{F}(P^{\textbf{q}}))$ \Comment{Patch Encoding}

\While{$m \leq M$}
    \State $F^{\textbf{q}}_m \leftarrow CA(MLP_{F}(\textbf{o}^{\textbf{q}}_{m}), F^{\textbf{q}}_{P})$ \Comment{Cross Attention at $\textbf{o}^{\textbf{q}}_{m}$}
    \State $[x^{\textbf{q}}_m \: y^{\textbf{q}}_m \: z^{\textbf{q}}_m]^T \leftarrow MLP_{I}(F^{\textbf{q}}_m)$ \Comment{Implicit nearest point}
    \State $t^{\textbf{q}}_m \leftarrow \left\|[x^{\textbf{q}}_m \: y^{\textbf{q}}_m \: z^{\textbf{q}}_m]\right\|_2$ \Comment{Implicit nearest distance}
    \State $\textbf{n}^{\textbf{q}}_m \leftarrow [x^{\textbf{q}}_m \: y^{\textbf{q}}_m \: z^{\textbf{q}}_m]^T / t^{\textbf{q}}_m$ \Comment{Implicit nearest direction}
    \State $\Tilde{t}^{\textbf{q}} = \Tilde{t}^{\textbf{q}} + t^{\textbf{q}}_m$ \Comment{Update cumulative depth}
    \State $m = m+1$ \Comment{Increment marching step index}
    \State $\textbf{o}^{\textbf{q}}_{m} \leftarrow \textbf{d}^{\textbf{q}} \cdot \Tilde{t}^{\textbf{q}}$ \Comment{Update origin}
\EndWhile
\State $F^{\textbf{q}}_m \leftarrow CA(MLP_{F}(\textbf{o}^{\textbf{q}}_{m}), F^{\textbf{q}}_{P})$ \Comment{Cross Attention at $\textbf{o}^{\textbf{q}}_{m}$}
\State $\epsilon^{\textbf{q}} \leftarrow MLP_{\epsilon}(concatenate(F^{\textbf{q}}_m, \textbf{d}^{\textbf{q}}))$ \Comment{Epsilon estimation} \\
\Return $\hat{t}^{\textbf{q}} \leftarrow \Tilde{t}^{\textbf{q}} + \epsilon^{\textbf{q}}$
\end{algorithmic}

\end{algorithm}

Following the sphere tracing paper \cite{sphere_tracing}, we define our ray mapping function, $\textbf{q}: \mathbb{R} \rightarrow \mathbb{R}^3$, as:
\begin{align*} \label{eq:ray_def}\tag{1} 
\textbf{q}(t) = \textbf{o}^{\textbf{q}} + t \circ \textbf{d}^{\textbf{q}},
\end{align*}
where $\textbf{o}^{\textbf{q}} \in \mathbb{R}^{3}$ is the origin and $\textbf{d}^{\textbf{q}} \in \mathbb{R}^{3}$ is the unit vector that defines the direction. The problem formulation of the method is similar to the work of Kwon et al. \cite{kwon2022implicit}:
\begin{align*} \label{eq:upsampling}\tag{2} 
U = S \cup \{\textbf{q}(\hat{t}^{\textbf{q}}) \; | \; \forall{\textbf{q}} \in Q , \; \hat{t}^{\textbf{q}} = f(P^{\textbf{q}}, \textbf{d}^{\textbf{q}})\},
\end{align*}
where $S\subset \mathbb{R}^{3}$ is the sparse point cloud, $U\subset \mathbb{R}^{3}$ the upsampled point cloud, $Q = \{\textbf{q} \; | \; \textbf{o}^{\textbf{q}}, \textbf{d}^{\textbf{q}} \in \mathbb{R}^{3}\}$ the set of query rays, $P^{\textbf{q}} \in \mathbb{R}^{3 \times k}$ the patch that corresponds to ray $\textbf{q}$, and $\hat{t}^{\textbf{q}} \in \mathbb{R}^{1}$ the predicted depth that corresponds to ray $\textbf{q}$. In our method, the number of query rays is set to $|Q| = |S|\cdot (r-1)$, where $r$ is the upsampling rate. The ray depth prediction network $f(\cdot)$ takes in $\textbf{d}^{\textbf{q}}$ and $P^{\textbf{q}}$, where the $k$-NN patch $P^{\textbf{q}} = [\textbf{p}^{\textbf{q}}_1 \,...\, \textbf{p}^{\textbf{q}}_i \,...\, \textbf{p}^{\textbf{q}}_k]$ is sampled from $S$, as described in Sect. \ref{sect:query-patch}, and translated for relative positioning so that $\textbf{o}^{\textbf{q}}$ is centered at $\textbf{o}^{\textbf{q}}_{m=1}=(0,0,0)$, where $m=1$ means the marching step index initialized to 1 (Alg. \ref{alg:ray_marching}). In other words, $\{\textbf{p}^{\textbf{q}}_i+\textbf{o}^{\textbf{q}}\}_{i=1}^k \subset S$. Relative positioning introduces simpler loss functions and prevents overfitting by ignoring the ray origin information during training. The depth of a ray, $\hat{t}^{\textbf{q}}$, indicates the distance from the query ray origin to the crossing point to the implicit surface (Fig. \ref{fig:ray_marching}).

\subsubsection{Neural} implicit surface definition with $MLP_I$
We define our implicit surface similar to that in SAPCU \cite{sapcu} with a UDF since a signed distance function is impossible to define in a patch-based method. A multi-layer perceptron (MLP), $MLP_I$ (Alg. \ref{alg:ray_marching}; Fig. \ref{fig:network}), following the feature encoding of $F^{\textbf{q}}_m$, outputs the nearest point, $[x^{\textbf{q}}_m \: y^{\textbf{q}}_m \: z^{\textbf{q}}_m]^T$, from the implicit surface to an arbitrary $\textbf{o}^{\textbf{q}}_m$. The implicit nearest points (middle of Figs. \ref{fig:intro} and \ref{fig:kitti}) is decomposed into meaningful information about the distance, $t^{\textbf{q}}_m = \left\| [x^{\textbf{q}}_m \: y^{\textbf{q}}_m \: z^{\textbf{q}}_m]\right\|_2$, and direction, $\textbf{n}^{\textbf{q}}_m = [x^{\textbf{q}}_m \: y^{\textbf{q}}_m \: z^{\textbf{q}}_m]^T / t^{\textbf{q}}_m$, from $\textbf{o}^{\textbf{q}}_m$ to $(x^{\textbf{q}}_m, y^{\textbf{q}}_m, z^{\textbf{q}}_m)$. 

Firstly, a unit vector, $\textbf{n}^{\textbf{q}}_m$, combined with $\textbf{o}^{\textbf{q}}_m$ defines a plane (The purple line in the inset of Fig. \ref{fig:ray_marching}), where each $\textbf{p}^{\textbf{q}}_i$ can be projected to with distance denoted as $proj^{\textbf{q}}_{m,i}$:
\begin{align*} \label{eq:proj}\tag{3}
&S_{cos}(\textbf{a}, \textbf{b}) = \frac{\textbf{a} \cdot \textbf{b}}{\left\|\textbf{a}\right\| \, \left\|\textbf{b}\right\|}, \\
&proj^{\textbf{q}}_{m, i} = S_{cos}\left(\textbf{n}^{\textbf{q}}_{m}, \frac{\textbf{o}^{\textbf{q}}_m \textbf{p}^{\textbf{q}}_i}{\left\|\textbf{o}^{\textbf{q}}_m \, \textbf{p}^{\textbf{q}}_i\right\|_2}\right) \cdot \left\|\textbf{o}^{\textbf{q}}_m \textbf{p}^{\textbf{q}}_i\right\|_2, 
\end{align*}
where $S_{cos}(\textbf{a}, \textbf{b})$ is the cosine similarity between two unit vectors. The cosine similarity between $\textbf{n}^{\textbf{q}}_m$ and the unit vector of $\textbf{o}^{\textbf{q}}_m \, \textbf{p}^{\textbf{q}}_i$ is multiplied by the magnitude between the two points, $\left\|\textbf{o}^{\textbf{q}}_m \, \textbf{p}^{\textbf{q}}_i\right\|_2$, to obtain the projection distance, $proj^{\textbf{q}}_{m,i}$, defined above (Inset of Fig. \ref{fig:ray_marching}). In the optimal situation, the plane defined with $\textbf{n}^{\textbf{q}}_m$ and $\textbf{o}^{\textbf{q}}_m$ would be parallel to the tangent plane that touches the object surface at $(x^{\textbf{q}}_m, y^{\textbf{q}}_m, z^{\textbf{q}}_m)$, which makes $\textbf{n}^{\textbf{q}}_m$ equivalent to the surface normal, namely, the implicit nearest direction. We approximate the tangent plane using linear regression on the projection distances of the points to the tangent plane. An optimal solution using $l2$ loss with the 1D distance data results in the mean $\mu^{\textbf{q}}_{m} = |\sum^{k}_{i=1} proj^{\textbf{q}}_{m, i}/k|$ as the offset between this tangent plane and the plane defined with $\textbf{n}^{\textbf{q}}_m$ and $\textbf{o}^{\textbf{q}}_m$ (Inset of Fig. \ref{fig:ray_marching}). Therefore, the training objective for the optimal $\textbf{n}^{\textbf{q}}_m$ is by minimizing the term $|proj^{\textbf{q}}_{m, i} - \mu^{\textbf{q}}_m|$. This term is weighed differently with $w_{m, i}$ based on the Euclidean distances between $\textbf{p}^{\textbf{q}}_i$'s and $\textbf{o}^{\textbf{q}}_m$ to focus more on adjacent points than on points further away on a complex surface:
\begin{align*}\label{eq:L_tan}\tag{4}
&\omega^{\textbf{q}}_{m, i} = \exp{\left(-\frac{\left\|\textbf{o}^{\textbf{q}}_m \, \textbf{p}^{\textbf{q}}_i\right\|_2^2}{2 \cdot avg(\{\left\|\textbf{o}^{\textbf{q}}_m \, \textbf{p}^{\textbf{q}}_j\right\|_2^2 \;|\; \forall \textbf{p}^{\textbf{q}}_j \in P^{\textbf{q}}\})}\right)},\\
&L_{tan}^{\textbf{q}} = \frac{1}{M} \cdot \sum\limits^{M}_{m=1} \sqrt{
    \frac 
    {\sum\limits^{k}_{i=1} [(proj^{\textbf{q}}_{m, i} - \mu^{\textbf{q}}_{m}) \cdot \omega^{\textbf{q}}_{m, i}]^2}
    {\sum\limits^{k}_{i=1}\omega^{\textbf{q}}_{m, i}}
},
\end{align*}
where $avg(\cdot)$ is the mean value of a set, and $M$ is the maximum number of marching steps.

The best approximation of the implicit nearest distance, $t^{\textbf{q}}_m$, is the projection distance of the nearest point in $P^{\textbf{q}}$ from $\textbf{o}^{\textbf{q}}_m$ due to limited information in the point set:
\begin{align*}\label{eq:L_ms}\tag{5}
&L_{ms}^{\textbf{q}} = \frac{1}{M} \cdot \sum\limits^{M}_{m=1}|t^{\textbf{q}}_m - proj^{\textbf{q}}_{m, \hat{i}} |, 
\end{align*}
where $\hat{i}=\argmin\limits_i \left\|\textbf{o}^{\textbf{q}}_m \, \textbf{p}^{\textbf{q}}_i\right\|_2$ is the index of the point $\textbf{p}^{\textbf{q}}_i$ that minimizes the Euclidean distance from the origin $\textbf{o}^{\textbf{q}}_m$. Cumulative query ray depth, $\Tilde{t}^{\textbf{q}} = \sum_{m=1}^{M} t^{\textbf{q}}_m$, is the sum of all $t^{\textbf{q}}_m$ through multiple marching steps (Alg. \ref{alg:ray_marching} and Fig. \ref{fig:network}). $\textbf{o}^{\textbf{q}}_m$ is updated to $\textbf{d}^{\textbf{q}} \cdot \Tilde{t}^{\textbf{q}}$ at every step when $\Tilde{t}^{\textbf{q}}$ is accumulated with $t^{\textbf{q}}_m$.

\subsubsection{Neural} implicit surface definition with $MLP_\epsilon$
Sphere tracing \cite{sphere_tracing} does not guarantee precise query ray depth ($\Tilde{t}^{\textbf{q}} \neq \hat{t}^{\textbf{q}}$) as illustrated in Fig. \ref{fig:ray_marching}. Therefore, we have another distance function, $MLP_\epsilon$, to add a small offset $\epsilon^{\textbf{q}}$ to get $\hat{t}^{\textbf{q}} = \Tilde{t}^{\textbf{q}} + \epsilon^{\textbf{q}}$ (Alg. \ref{alg:ray_marching}; Fig. \ref{fig:network}). The difference in the implicit surface definition is that the directional unit vector is given to $MLP_\epsilon$ as an input, and the function outputs the distance $\epsilon^{\textbf{q}}$ only. It enables querying in an arbitrary direction $\textbf{d}^{\textbf{q}}$ from an arbitrary $\textbf{o}^{\textbf{q}}_m$, which is essential for our user-controlled upsampling. The $\epsilon^{\textbf{q}}$ has a constraint to have a non-negative value following \cite{sphere_tracing}, which is represented by the loss function:
\begin{align*} \label{eq:L_epsilon}\tag{6} 
L_{\epsilon}^{\textbf{q}} = max(0, -\epsilon^{\textbf{q}}).
\end{align*}
This loss function penalizes the depth overestimation of the last epsilon value instead of placing the ReLU activation at the network's end (see Supplementary Material).
\subsubsection{Training objective}
Given the target depth $t^{\textbf{q}} \in \mathbb{R}^3$ of a training query ray, $\textbf{q}$, the training objective of $f(\cdot)$ is by minimizing the following:
\begin{align*} \label{eq:L_total}\tag{7}
&L_{MAE} = \frac{\sum^{Q}_{\textbf{q}} |\hat{t}^{\textbf{q}} - t^{\textbf{q}}|}{|Q|}, \;\; 
L_{RMSE} = \sqrt{\frac{\sum^{Q}_{\textbf{q}} (\hat{t}^{\textbf{q}}-t^{\textbf{q}})^2}{|Q|}},  \\
&L_{total} = L_{MAE} + L_{RMSE} + \omega_{ms} \cdot L_{ms} + \omega_{tan} \cdot L_{tan} + L_{\epsilon},
\end{align*}
where $0 \leq \omega_{ms} \leq 1$ and $0 \leq \omega_{tan} \leq 1$. The losses calculated per query ray, $L^{\textbf{q}}_{tan}$, $L^{\textbf{q}}_{ms}$ and $L^{\textbf{q}}_{\epsilon}$ are averaged (e.g. $L_{tan } = \sum^{Q}_\textbf{q} L_{tan}^{\textbf{q}} / |Q|$) to obtain the final loss. 

\subsection{Query ray and patch generation}
\label{sect:query-patch}
The origin $\textbf{o} \in O$ of query rays is sampled using surface normal estimation for each $k$ nearest neighbourhood ($k$-NN; $k=16$), $\EuScript{N} \in\mathbb{R}^{k \times 3} $, of downsampled points in the input point cloud:
\begin{align*} \label{eq:origins}\tag{8} 
\textbf{o} = \mu_{\EuScript{N}} \pm (\lbrack\lambda_x \; \lambda_y \; \lambda_z \rbrack ^T )^{\circ\frac12},
\end{align*}
where $\mu_{\EuScript{N}} \in \mathbb{R}^{3}$ is the mean point of $\EuScript{N}$ and $\lbrack\lambda_x \; \lambda_y \; \lambda_z \rbrack^T$ is the vector of eigenvalues of the covariance matrix $(\EuScript{N}-\mu_{\EuScript{N}})^T \cdot (\EuScript{N}-\mu_{\EuScript{N}})$. Because elements are squared or multiplied by each other for covariance calculations, Hadamard square root, $\lbrack\cdot\rbrack^{\circ\frac12}$, is applied to the eigenvalues. The $\pm$ sign is randomly selected with a binomial distribution. 

The generation of query rays and patches is based on arbitrary query points, which can be known points in existing point clouds or a novel point from a rule-based method. The query ray origin corresponding to each query point is determined by searching the nearest $\textbf{o}$. Query ray directions $\textbf{d}^{\textbf{q}}$ are defined as the unit vector from a query ray origin to query points. Given a query point, $k$-NN patches ($k=16$), $P^{\textbf{q}}$, are sampled from $S$ followed by relative positioning. Random rotations are applied to patches to augment the data. Depths are normalized to the $(0, 1]$ range by scaling, where the maximum depth between $\textbf{o}$ and $P^{\textbf{q}}$ is set as 1.

\subsubsection{Known query point sampling for supervised and self-supervised learning}
Since the target depths are required to train the proposed network, the known query points and the corresponding depths in the dense ground truth are used for supervised learning. Self-supervised learning is achieved by replacing the query points of dense ground truth point clouds with sparse input point clouds.

\subsubsection{Novel query point generation for upsampling}
\label{sect:novel_query}
We propose a mid-point algorithm similar to that of Grad-PU \cite{gradpu}. In particular, for every point $\textbf{s} \in S$, the method finds six nearest points, calculates the midpoint from $\textbf{s}$ to each of these neighbours and takes each as a query point. We employ an additional constraint where the angles between the vectors from $\textbf{s}$ to midpoints in the neighbourhood, $\EuScript{N}^{\textbf{s}}_{syn}$, should not be less than $\pi / 6$ to maintain a hexagonal shape for uniformity, following the observation by Li et al. \cite{pu_gan}. To accomplish this, we give indices to neighbouring points of $s$ from the nearest to the farthest, $S' = \{\textbf{s}'_1,...,\textbf{s}'_i,...\}$. Query points, $\Tilde{Q}_{syn}$, are:
\begin{align*} \label{eq:syn_query_sampling}\tag{8} 
&\EuScript{N}^{\textbf{s}}_{syn} = \bigcup\limits_{i=1}^n \{ \textbf{s}'_i | \; \forall{j} \leq i ,\; \angle \textbf{s}'_i \, \textbf{s} \, \textbf{s}'_j \; < \frac{\pi}{6} \}, \\
&\Tilde{Q}_{syn} = \bigcup\limits_\textbf{s}^S \{ \frac{s + s'}{2} | \; \forall{s'} \in \EuScript{N}^{\textbf{s}}_{syn}\},
\end{align*}
where $n$ is the smallest number that satisfies $|\EuScript{N}^{\textbf{s}}_{syn}| = 6$.

We adapt our rule-based query generation method to real-scanned point clouds by selecting points, $\{\textbf{s}''_1,...,\textbf{s}''_i,...,\textbf{s}''_{n=8}\}$, with the eight smallest $\angle \textbf{s} \boldsymbol{\alpha} \textbf{s}''_i$, where $ \boldsymbol{\alpha}$ is the sensor location. We define the neighbourhood of $s$ as
\begin{align*} \label{eq:real_neighbour}\tag{9} 
&\EuScript{N}^{\textbf{s}}_{real} = \{ \textbf{s}''| \; \forall{\textbf{s}''} \in S'' , \; d_{near} < \left\| \textbf{s} \, \textbf{s}'' \right\|_2^2 < d_{far} \}, 
\end{align*}
where $d_{near}$ and $d_{far}$ are, respectively the second nearest and the second farthest distances from $\textbf{s}$ to $S''$ to reject outliers. This adaptation is more robust to the real-scanned data since LiDAR point clouds are more uniformly distributed in spherical coordinates than in Cartesian coordinates. We have the following constraint to upsample in sparse regions:
\begin{align*} \label{eq:real_query_sampling}\tag{10}
&V = \bigcup\limits_\textbf{s}^S \{ \textbf{s} \, \textbf{s}''\; | \; \forall{\textbf{s}''} \in \EuScript{N}^{\textbf{s}}_{real} \}, \\
&\hat{V} = \{ \textbf{v} | \; \forall{\textbf{v}} \in V, \, \left\| \textbf{v} \right\|_2^2 > median(\{ \left\| \textbf{v} \right\|_2^2 \; | \; \forall{\textbf{v}} \in V \})\}, \\
&\Tilde{Q} = \bigcup\limits_{n=1}^{N} \{ \textbf{s}^{\hat{\textbf{v}}} + t \circ \textbf{d}^{\hat{\textbf{v}}} | \; \forall{\hat{\textbf{v}}} \in \hat{V}, \, t = \frac{n}{N + 1} \cdot \left\| \hat{\textbf{v}} \right\|_2^2 \},
\end{align*}
where $\Tilde{Q}$ is the set of query points and $N = \lceil |\hat{V}| / (|S| \cdot r) \rceil$. Points in the upsampled point clouds that are outliers concerning their local neighbourhoods in input point clouds are omitted. This defines the ROI of our upsampling objective.

\subsection{Network architecture}
\begin{figure*}[!t]
\centering
\includegraphics[width=\textwidth]{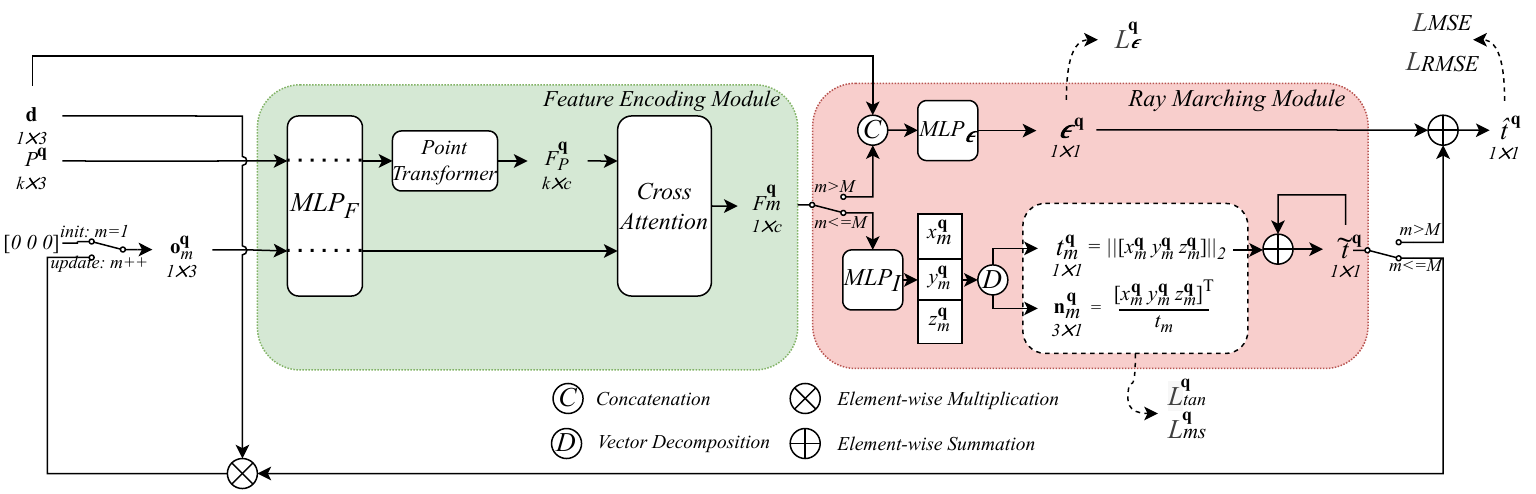}
\caption{Network overview of PU-Ray with a single query input defined with $\textbf{d}^{\textbf{q}}$ and $P^{\textbf{q}}$.}
\label{fig:network}
\end{figure*}

\label{sect:network}
Our architecture consists of feature encoding and ray marching modules (Figure \ref{fig:network}). The feature encoding module transforms the 3D coordinates in a patch, $P^{\textbf{q}}$, into a higher dimension space incorporating neighbourhood information. The encoded features are accumulated in the ray marching module to define the UDF of $P^{\textbf{q}}$.
\subsubsection{Feature encoding module}
\label{sect:feature}
We adopt the Point Transformer's (PT) self-attention module \cite{point_transformer} to encode the points within patch $P^{\textbf{q}}$. First, the points in $P^{\textbf{q}}$ are fed into a multi-layer perceptron, $MLP_F$, shared throughout the entire network, to extract $c=32$ features from the $(x, y, z)$ coordinates of $\textbf{p}^{\textbf{q}}_i$, which are inputs to three different layers for key, query, and value in the Point Transformer. The relative coordinates between points are used for positional encoding as in the Point Transformer \cite{point_transformer}. The output transformed features, $F^{\textbf{q}}_P$, have inter-point spatial relationship information with other samples in the patch, making it advantageous for representation learning of neural implicit surfaces. To obtain $F^{\textbf{q}}_m$ in Alg. \ref{alg:ray_marching} and Fig. \ref{fig:network}, the objective is to calculate the cross-attention (CA) of $\textbf{o}^{\textbf{q}}_m$ from $F^{\textbf{q}}_{P}$ using the ablated architecture of the Point Transformer's attention module \cite{point_transformer}. $F^{\textbf{q}}_{m}$ is fed into three different linear layers to obtain the key and the value, while $\textbf{o}^{\textbf{q}}_m$ is directly used for the query. The relative 3D coordinates of the local neighbourhood, $P^{\textbf{q}}$, centred at $\textbf{o}^{\textbf{q}}_m$, are input to a separate linear layer for position encoding. 
\subsubsection{Ray marching module}
\label{sect:ray_marching_module}
The ray marching module contains $MLP_I$ and $MLP_\epsilon$ that define the neural implicit surface. Depending on the MLP that follows the cross-attention module, two types of 3D coordinates are retrieved given the encoded features $F^{\textbf{q}}_m$ (Fig. \ref{fig:network}): 1) $MLP_I$ produces the approximate nearest point, $(x^{\textbf{q}}_m, y^{\textbf{q}}_m, z^{\textbf{q}}_m)$, on the implicit surface from $\textbf{o}^{\textbf{q}}_m$, or 2) $MLP_\epsilon$ produces the final $\epsilon^{\textbf{q}}$ value from $\textbf{o}^{\textbf{q}}_m$ to the point where the query ray hits the surface. Both MLPs consist of three fully connected layers with a size of 32, followed by the ReLU activation. The output layers are of size of 3 and 1 are added to the end of $MLP_I$ and $MLP_\epsilon$, respectively.

\section{Experiments}
\subsection{Experimentation setups}

Following the benchmarks from existing studies, the PU-GAN \cite{pu_gan} and PU1K \cite{pu_gcn} datasets are used for quantitative and qualitative experimentations. The PU-GAN \cite{pu_gan} dataset has 120 and 27 training and testing mesh models, respectively. Since no point clouds are provided, the Poisson disk sampling method \cite{poisson} is used to obtain the input point cloud and the ground truth samples using the program provided by He et al. \cite{gradpu}. PU1K has a larger testing dataset with 127 point clouds. We train our models only on the PU-GAN dataset to analyze the domain dependency. The training dataset has simple, medium and complex sub-datasets, each with 40 shapes. 

Moreover, we evaluate our method on the 3D reconstruction task for ITS \cite{completion}, with a simulated real-scanned dataset using the Vista simulator's \cite{vista} occlusion removal on densely accumulated point clouds. KITTI-360 \cite{kitti360} (41 point clouds) and private data collected on highways (233 point clouds) are used for comprehensive assessments in different road environments. The input, low-resolution real-scanned simulation data, follows the configuration settings of the KITTI-360's HDL-64E sensor \cite{kitti360, velodyne}. Experiments are in the far range with sparse point density ($>15m$ and $>30m$ from the sensor for urban and highway environments). Input far-range point clouds have 9741.80 and 15592.06 points on average for urban and highway environments, respectively. We also simulate 2$\times$ super-resolution in the horizontal and vertical axes of the range image, similar to that of the SOTA Velodyne Alpha Prime VLS-128 sensor \cite{lidar_performance}. The reconstruction ground truths are defined by omitting points in the accumulated point clouds outside of the ROI, as in \ref{sect:novel_query}, followed by FPS. Outliers in the super-resolution point clouds are omitted to mitigate unfair over-penalization. Snippets of KITTI-360 \cite{kitti360} scenes are used to assess the application on real-scanned data qualitatively.

Six models are trained: \emph{Supervised}, \emph{Self}$_{Test}$, \emph{Self}$_{Train}$, \emph{Self}$_{Simple}$, \emph{Self}$_{Medium}$ and \emph{Self}$_{Complex}$. \emph{Supervised} is the only supervised model where the query point set is extracted from the dense ground truth point clouds. \emph{Self}$_{Test}$ and \emph{Self}$_{Train}$ are self-supervised models, with the input point clouds from the training and test datasets, respectively. \emph{Self}$_{Simple}$, \emph{Self}$_{Medium}$ and \emph{Self}$_{Complex}$ are self-supervised models like the previous two but use a single point cloud with 2,048 training points, randomly sampled from the single, medium and complex training datasets, respectively. The weights of the loss functions are set as $\omega_{ms}, \omega_{tan}=0.1$ for supervised training, and $\omega_{ms}, \omega_{tan}=0.5$ for self-supervised learning. Self-supervised learning tends to overfit to the final depth by using input points for both query and patch generation. Thus, the weights on the arbitrary implicit surface learning using $L_{tan}$ and $L_{ms}$ compared to the precise final depth estimation using $L_{MSE}$, $L_{RMSE}$ and $L_{\epsilon}$ are relatively higher for self-supervised training than for supervised training. Different numbers of epochs are also used because of the overfitting problem (\emph{Supervised}: 100, \emph{Self}$_{Test}$, \emph{Self}$_{Train}$: 15, \emph{Self}$_{Simple}$: 30, \emph{Self}$_{Medium}$: 30 and \emph{Self}$_{Complex}$: 30). The batch size is set as $\lfloor |Q|/64 \rfloor$. The initial learning rate is set as 0.005 and decayed with a rate of 0.99 every epoch with the Adam optimizer \cite{adam}. The models are trained on an NVIDIA RTX 3080 GPU Light Hash Rate (LHR) with 10GB of memory, except for the supervised model, which was trained on the NVIDIA RTX A6000 GPU with 48GB to accommodate the larger dataset. The size of query ray origins per sample is $|O|=128$ to have diverse angles between the tangent plane and the query vector. The choice of the inference $|O|$ is reasoned by the $k$-NN size of 16 for the query ray origin generation (e.g. $128 = 2048/16$). Thus, a query ray origin covers a similar neighbourhood area generated in the inference stage. The number of marching steps is set to 6, determined based on an ablation study.

\subsection{Applications to Synthetic Datasets}

\begin{table}[!htb]
\caption{Quantitative results on the PU1K test dataset. All metric units are multiplied by a factor of $10^{-3}$. Italic labels indicate a self-supervised method. The best three results are coloured red ($1^{st}$), orange ($2^{nd}$) and green ($3^{rd}$).}
\label{table:pu1k_results}
\let\center\empty
\let\endcenter\relax
\centering
\begin{tabular}{lccc}
\hline
\multicolumn{1}{c}{Methods}                & CD $\downarrow$ & HD $\downarrow$ & P2F $\downarrow$ \\ \hline
PU-Net\cite{pu_net}                        & 1.155 & 15.170 & 4.834 \\
MPU\cite{mpu}                              & 0.935 & 13.327 & 3.511 \\
PU-GACNet\cite{pugac}                      & 0.665 & 9.053  & 2.429 \\
PU-GCN\cite{pu_gcn}                        & 0.585 & 7.577  & 2.499 \\
PU-Transformer\cite{pu_transformer}        & 0.451 & \textcolor{orange}{3.843}  & \textbf{\textcolor{red}{1.277}} \\
Grad-PU\cite{gradpu}                       & \textbf{\textcolor{red}{0.404}} & \textbf{\textcolor{red}{3.732}}  & \textcolor{green}{1.474} \\ \hline
\multicolumn{1}{c}{Ours} & \multicolumn{3}{c}{} \\ \hline
Supervised                                  & \textbf{\textcolor{red}{0.404}} & \textcolor{green}{4.025}  & \textcolor{orange}{1.341} \\
\emph{Self}$_{train}$                                  & 0.431 & 4.537  & 1.820 \\
\emph{Self}$_{test}$                                   & \textcolor{green}{0.427} & 4.705  & 1.857 \\
\emph{Self}$_{simple}$                                 & 0.576 & 12.963 & 2.114 \\
\emph{Self}$_{medium}$                                 & 0.444 & 5.181  & 2.024 \\
\emph{Self}$_{complex}$                                 & 0.452 & 5.213  & 2.160 \\ \hline
\end{tabular}

\end{table}

\begin{table}[!htb]
\caption{Quantitative results on the PU-GAN test dataset. All metric units are multiplied by a factor of $10^{-3}$. Italic labels indicate a self-supervised method. The best three results are coloured red ($1^{st}$), orange ($2^{nd}$) and green ($3^{rd}$).}
\label{table:pugan_results}
\let\center\empty
\let\endcenter\relax
\centering
\begin{center}
\begin{tabular}{l|ccc|ccc}
\hline
\multicolumn{1}{r|}{} & \multicolumn{3}{c|}{$r = 4$} & \multicolumn{3}{c}{$r = 16$} \\ \cline{2-7} 
\multicolumn{1}{c|}{Methods}  & CD $\downarrow$ & HD $\downarrow$ & P2F $\downarrow$ & CD $\downarrow$ & HD $\downarrow$ & P2F $\downarrow$ \\ \hline
PU-Net\cite{pu_net}  & 0.493 & 4.508 & 4.315 & 0.510 & 6.739 & 5.442 \\
MPU\cite{mpu}    & 0.305 & 4.463 & 2.882 & 0.187 & 6.243 & 3.183 \\
PU-GAN\cite{pu_gan} & 0.296 & 5.722 & 2.812 & 0.229 & 7.653 & 3.304 \\
PU-GCN\cite{pu_gcn} & 0.291 & 2.986 & \textcolor{green}{2.471} & 0.158 & 3.774 & \textcolor{green}{2.592} \\
\emph{SAPCU}\cite{sapcu}  & 0.465 & 10.572 & 3.421 & 0.510 & 6.739 & 5.442 \\
Grad-PU\cite{gradpu} & \textcolor{orange}{0.268} & \textbf{\textcolor{red}{2.601}} & \textcolor{orange}{1.990} & \textcolor{orange}{0.126} & \textbf{\textcolor{red}{2.628}} & \textcolor{orange}{2.233} \\ \hline
\multicolumn{1}{c|}{Ours} & \multicolumn{3}{c|}{} & \multicolumn{3}{c}{} \\ \hline
Supervised          & \textbf{\textcolor{red}{0.260}} & \textcolor{orange}{2.736} & \textbf{\textcolor{red}{1.707}} & \textbf{\textcolor{red}{0.121}} & \textcolor{orange}{2.654} & \textbf{\textcolor{red}{2.093}} \\ 
\emph{Self}$_{train}$     & \textcolor{green}{0.274} & \textcolor{green}{2.797} & 2.721 & 0.155 & 2.970 & 3.384 \\
\emph{Self}$_{test}$      & 0.293 & 2.864 & 2.756 & \textcolor{green}{0.141} & \textcolor{green}{2.694} & 3.355 \\
\emph{Self}$_{simple}$    & 0.290 & 6.030 & 3.109 & 0.161 & 5.335 & 3.757 \\
\emph{Self}$_{medium}$    & 0.301 & 3.260 & 3.093 & 0.160 & 3.200 & 3.680 \\
\emph{Self}$_{complex}$   & 0.325 & 3.400 & 3.427 & 0.181 & 3.304 & 4.355 \\ \hline
\end{tabular}
\end{center}

\end{table}

\subsubsection{Results on PU1K and PU-GAN datasets}
Since PU1K's point cloud inputs are fixed, all the results of the compared methods are taken from the existing papers. Table \ref{table:pu1k_results} shows that our approach with supervised learning, even with training using the smaller PU-GAN dataset, performs at the SOTA level along with PU-Transformer \cite{pu_transformer} and Grad-PU \cite{gradpu}. The Chamfer Distance result, the best of all compared methods and tied with Grad-PU, is achieved by training without any loss functions for reducing the metric itself or its similarly designed alternatives. Our Hausdorff and P2F distances are ranked in the third and second places, respectively, with small margins compared to SOTA methods. The training objective is mainly on depth prediction, and the naive rule-based query generation method determines the uniform distribution in the output. This illustrates that uniform distribution in the output can be achieved using a simple method instead of a model with a vast number of network parameters to fit the ground truth. A similar trend of PU1K results is also observed in the PU-GAN dataset results (Table \ref{table:pugan_results}). Our supervised method achieves the lowest Chamfer and P2F distances and the second-best Hausdorff distances of all compared methods in $4\times$ and $16\times$ upsampling. The significant margins in the P2F differences between our method and the Grad-PU \cite{gradpu} illustrate the importance of accurate depth prediction of a query ray. Other fixed-rate methods \cite{pu_net, mpu, pu_gan, pu_gcn}, and upsample $4\times$ twice for $16\times$ upsampling, add more noise to the output. On the other hand, PU-Ray introduces less noise using an arbitrary scaling rate, $r$. The visual quality of our results is organized in the Supplementary Material. The self-supervised models also show promising results compared to some existing methods PU-Net \cite{pu_net}, MPU \cite{mpu}, PU-GACNet \cite{pugac} and PUGCN \cite{pu_gcn} (Table \ref{table:pugan_results} and Supplementary Material). Moreover, the models trained on a single point-cloud input demonstrate that our method does not require a large dataset to achieve a significant performance improvement, demonstrating its potential for zero-shot learning. All of our five self-supervised models generally outperform SAPCU \cite{sapcu}. Accurate ground-truth depths could provide better guidance for training, while SAPCU has to rely on manually defined implicit surface representations.

\subsubsection{Robustness to challenging inputs}
\begin{table}[!htb]
\caption{Quantitative results on the noisy PU-GAN test dataset. All metric units are multiplied by a factor of $10^{-3}$. Italic labels indicate a self-supervised method. The best three results are coloured red ($1^{st}$), orange ($2^{nd}$) and green ($3^{rd}$).}
\label{table:noise_results}
\let\center\empty
\let\endcenter\relax
\centering
\begin{center}
\begin{tabular}{l|ccc|ccc}
\hline
\multicolumn{1}{r|}{} & \multicolumn{3}{c|}{$\gamma = 0.01$} & \multicolumn{3}{c}{$\gamma = 0.02$} \\ \cline{2-7} 
\multicolumn{1}{c|}{Methods}  & CD $\downarrow$ & HD $\downarrow$ & P2F $\downarrow$ & CD $\downarrow$ & HD $\downarrow$ & P2F $\downarrow$ \\ \hline
PU-Net\cite{pu_net}    & 0.606 & 6.291 & 9.748 & 1.010 & 10.449 & 16.156 \\
MPU\cite{mpu}          & 0.460 & 8.692 & 7.262 & 0.782 & 10.264 & 13.589 \\
PU-GAN\cite{pu_gan}    & 0.457 & 6.056 & 7.294 & 0.795 & 9.089 & 14.217 \\
PU-GCN\cite{pu_gcn}    & \textbf{\textcolor{red}{0.436}} & 5.596 & \textcolor{green}{6.981} & 0.784 & 8.687 & 13.583 \\
\emph{SAPCU}\cite{sapcu}  & 0.728 & 16.707 & 10.933 & \textbf{\textcolor{red}{0.731}} & 16.829 & 10.933 \\
Grad-PU\cite{gradpu}   & \textcolor{orange}{0.448} & \textbf{\textcolor{red}{4.474}} & \textbf{\textcolor{red}{6.447}} & 0.767 & \textcolor{orange}{7.194} & \textcolor{orange}{11.417} \\ \hline
\multicolumn{1}{c|}{Ours} & \multicolumn{3}{c|}{} & \multicolumn{3}{c}{} \\ \hline
Supervised              & \textcolor{green}{0.454} & 4.745 & \textcolor{orange}{6.872} & 0.749 & \textbf{\textcolor{red}{7.185}} & 11.897\\
\emph{Self}$_{train}$         & 0.466 & 4.794 & 7.390 & 0.749 & 7.441 & 12.056 \\
\emph{Self}$_{test}$          & 0.471 & \textcolor{orange}{4.664} & 7.221 & \textcolor{orange}{0.735} & 7.353 & 11.700 \\
\emph{Self}$_{simple}$        & 0.482 & 5.815 & 7.190 & 0.785 & 10.830 & \textbf{\textcolor{red}{11.270}} \\
\emph{Self}$_{medium}$        & 0.477 & \textcolor{green}{4.714} & 7.253 & \textcolor{green}{0.738} & \textcolor{green}{7.262} & \textcolor{green}{11.435} \\
\emph{Self}$_{complex}$       & 0.488 & 4.761 & 7.466 & 0.745 & 7.275 & 11.526 \\ \hline
\end{tabular}
\end{center}

\end{table}

\begin{table}[!htb]
\caption{Quantitative results of supervised methods on the PU-GAN test dataset with smaller input sizes. All metric units are multiplied by a factor of $10^{-3}$. The best three results are coloured red ($1^{st}$), orange ($2^{nd}$) and green ($3^{rd}$).}
\label{table:diff_size}
\let\center\empty
\let\endcenter\relax
\centering
\begin{center}
\begin{tabular}{l|cc|cc|cc}
\hline
\multicolumn{1}{r|}{} & \multicolumn{2}{c|}{$|S|=256$} & \multicolumn{2}{c|}{$|S|=512$} & \multicolumn{2}{c}{$|S|=1024$} \\ \cline{2-7} 
\multicolumn{1}{c|}{Methods}  & CD $\downarrow$ & P2F $\downarrow$ & CD $\downarrow$ & P2F $\downarrow$ & CD $\downarrow$ & P2F $\downarrow$ \\ \hline
PU-Net\cite{pu_net}         & 3.073 & 17.044 & 1.719 & 11.657 & 0.923 & 7.020 \\
MPU\cite{mpu}               & \textcolor{green}{2.259} & 9.466 & 1.186 & 7.432 & 0.610 & 4.604 \\
PU-GAN\cite{pu_gan}         & \textbf{\textcolor{red}{2.152}} & 9.606 & 1.133 & 7.207 & \textbf{\textcolor{red}{0.572}} & 4.430 \\
PU-GCN\cite{pu_gcn}         & \textcolor{orange}{2.222} & \textcolor{green}{8.653} & \textcolor{green}{1.124} & \textcolor{green}{6.494} & \textcolor{green}{0.577} & \textcolor{green}{4.007} \\
Grad-PU\cite{gradpu}        & 2.266 & \textcolor{orange}{8.583} & \textbf{\textcolor{red}{1.075}} & \textcolor{orange}{5.495} & \textcolor{green}{0.577} & \textcolor{orange}{3.336} \\ \hline
Ours                        & 2.293 & \textbf{\textcolor{red}{8.281}} & \textcolor{orange}{1.105} & \textbf{\textcolor{red}{4.847}} & \textcolor{orange}{0.574} & \textbf{\textcolor{red}{2.841}} \\ \hline
\end{tabular}
\end{center}

\end{table}

PU-Ray's robustness to noise was tested by applying the method to perturbed input with Gaussian noise multiplied by two different factors, $\gamma=0.01$ and $\gamma=0.02$. While the three metric values are slightly behind Grad-PU \cite{gradpu} for $\gamma=0.01$, the best performances are achieved by several of our models learning at $\gamma=0.02$ (Table \ref{table:noise_results}). An observation suggests that our self-supervised model's performance degradation is less sensitive to a higher noise level than other methods. Also, our method can preserve the shape of the surface better than other methods (Supplementary Material). Evaluation with smaller input sizes shows how well the method adapts to point clouds with a different point density from the training datasets. Table \ref{table:diff_size} shows that our method consistently performs the best in the P2F distance metric and competitive results in the CD metric. It is noteworthy that CD is not a good evaluation metric when the compared point clouds have sparse point densities \cite{DCD}. HD metric is disregarded for the same reason. The results show the method's potential in real-scanned applications where point densities vary in different regions (i.e. decreasing density with the increasing distance to the sensor \cite{ultralidar}).

\subsection{Applications to real-scanned data}
\begin{figure*}[!t]
\centering
\includegraphics[width=\textwidth]{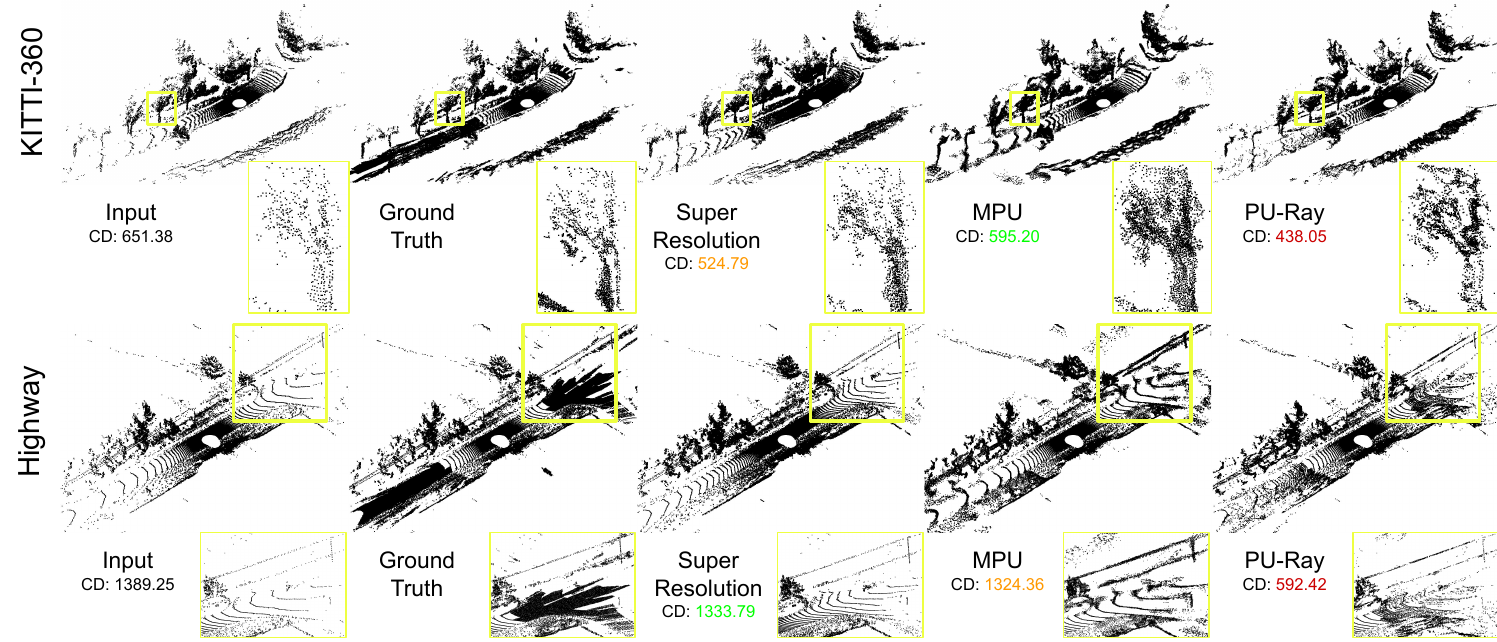}
\caption{Visual comparisons between the input point clouds, super-resolution \cite{vista}, MPU \cite{mpu}} upsampling and PU-Ray (Supervised) upsampling outputs of KITTI-360 \cite {kitti360} and highway datasets.
\label{fig:real_scanned}
\end{figure*}

\begin{table}[!htb]
\caption{Reconstruction results on real-scanned data. The best three results are coloured red ($1^{st}$), orange ($2^{nd}$) and green ($3^{rd}$).}
\label{table:real_scanned_results}
\let\center\empty
\let\endcenter\relax
\centering

\begin{center}
\begin{tabular}{l|cc|cc}
\hline
\multicolumn{1}{r|}{} & \multicolumn{2}{c|}{KITTI-360 ($>15m$)}                 & \multicolumn{2}{c}{Highway ($>30m$)} \\ \cline{2-5}
\multicolumn{1}{c|}{Methods}    &Avg. $r$ & CD $\downarrow$                     &Avg. $r$ & CD $\downarrow$ \\ \hline
Input (low-res) \cite{vista}    & - & 725.68                                    & - & 1295.29 \\
Super-res\cite{vista}           &$3.25 \times$ & \textcolor{orange}{541.59}     & $3.50 \times$ & 897.97 \\
PU-Net\cite{pu_net}             &4$\times$ & 807.99                             &4$\times$ & 1156.78 \\
MPU\cite{mpu}                   &4$\times$ & 729.01                             &4$\times$ & 1146.67 \\
PU-GAN\cite{pu_gan}             &4$\times$ & 776.20                             &4$\times$ & 1180.13 \\
PU-GCN\cite{pu_gcn}             &4$\times$ & 732.28                             &4$\times$ & 1164.67 \\
Grad-PU\cite{gradpu}            &4$\times$ & 753.16                             &4$\times$ & 1307.14 \\ \hline
\multicolumn{1}{c|}{Ours} & \multicolumn{2}{c|}{} & \multicolumn{2}{c}{} \\ \hline
Supervised                      &4$\times$ & \textbf{\textcolor{red}{523.82}}   &4$\times$ & \textbf{\textcolor{red}{788.99}}  \\
\emph{Self}$_{train}$           &4$\times$ & \textcolor{green}{555.88}          &4$\times$ & \textcolor{orange}{822.78} \\
\emph{Self}$_{test}$            &4$\times$ & 560.10                             &4$\times$ & 826.58 \\
\emph{Self}$_{simple}$          &4$\times$ & 595.18                             &4$\times$ & 838.91 \\
\emph{Self}$_{medium}$          &4$\times$ & 558.65                             &4$\times$ & \textcolor{green}{826.15} \\
\emph{Self}$_{complex}$         &4$\times$ & 567.13                             &4$\times$ & 831.25 \\ 
\hline

\end{tabular}
\end{center}

\end{table}

We demonstrate the effectiveness of our method in the 3D reconstruction task, which is an emphasized application of point cloud reinforcement for intelligent transportation systems (ITS) \cite{completion}. Quantitative (Table \ref{table:real_scanned_results}) and qualitative results (Figure \ref{fig:real_scanned}) show that the main strength of PU-Ray is demonstrated when it is applied to real-world scenes in both urban (KITTI-360 \cite{kitti360}) and highway environments. The consistently lower CDs suggest the robustness of PU-Ray compared to range-image super-resolution results in 3D reconstruction (Table \ref{table:real_scanned_results}). The upsampling behaviour of super-resolution in the 2D space fails to generate points with a uniform distribution in 3D, resulting in clustering (e.g. ring artifacts) (Fig. \ref{fig:real_scanned}). Occlusions by complex surfaces (e.g., trees) may have affected super-resolution to have a lower upsampling rate than $r=4$ in the far range, and instead, more points are added in the near range. Therefore, range-image super-resolution is not suitable for reconstruction, as it does not have the control of point generations in ROI. Our observation suggests that previous upsampling methods \cite{pu_net, mpu, pu_gan, pu_gcn, gradpu} do not overcome the domain dependency problem by generating points coherently with the input points (Fig. \ref{fig:real_scanned}). Compared to Grad-PU's \cite{gradpu} mid-point algorithm, our rule-based query generation allows points in the ROIs where there are wide gaps between input points, which is better suited for uneven distributions commonly observed in real-scanned data. Our method preserves the details, such as overhead power lines and tree branches, compared to the noisy output of MPU \cite{mpu} while having denser points than that of the super-resolution results (Fig. \ref{fig:real_scanned}). Even a patch-based method, MPU \cite{mpu}, could not generate implicit surfaces when data from unseen domains are provided. Illustrations in Fig. \ref{fig:kitti} and the Supplementary Material demonstrate upsampling on snippets of the KITTI-360 \cite{kitti360} dataset show that our method can upsample not only at the infrastructure level but also at the object level, such as pedestrians and cars.

\subsection{Computation Efficiency Assessment}
\begin{table}[!htb]
\caption{Comparison study of computation costs of upsampling on large-scale real-scanned datasets. A superscript indicates the GPU device used ($1080$: GeForce GTX 1080 Ti, $2080$: GeForce RTX 2080 Ti and $3080LHR$: GeForce RTX 3080 LHR).}
\label{table:computation}
\let\center\empty
\let\endcenter\relax
\centering
\begin{center}
\begin{tabular}{@{}l@{} | @{}c@{} | @{}c@{} | c c}
\hline
\multicolumn{1}{r|}{}  & \multicolumn{1}{c|}{Size} & \multicolumn{1}{c|}{Train} & \multicolumn{2}{c}{Inference ($sec / sample \downarrow$)} \\ \cline{4-5}
\multicolumn{1}{c|}{Methods}             & ($kb \downarrow$) & ($sec / epoch \downarrow$) & KITTI-360 & Highway \\ \hline
Grad-PU$^{1080}$\cite{gradpu}            & 401.8 & 262.00 & \textbf{3.70} & 10.69 \\
Ours$^{1080}$                           & \textbf{326.6} & \textbf{64.62} & 6.02 & \textbf{9.66} \\ \hline

PU-Net$^{2080}$\cite{pu_net}             & 796.7 & 71.41 & 5.87 & 12.38 \\
MPU$^{2080}$\cite{mpu}                   & 682.6 & 149.03 & 6.03 & 12.49 \\
PU-GAN$^{2080}$\cite{pu_gan}             & 1053.8 & 98.20 & 6.74 & 13.36 \\
PU-GCN$^{2080}$\cite{pu_gcn}             & 918.4 & 62.31 & 6.24 & 12.76 \\
Ours$^{2080}$                           & \textbf{326.6} & \textbf{58.15} & \textbf{5.33} & \textbf{8.22} \\ \hline

Ours$^{3080LHR}$                           & \textbf{326.6} & \textbf{35.33} & \textbf{3.13} & \textbf{4.97} \\ \hline
\end{tabular}
\end{center}

\end{table}

Our PU-Ray method achieves the smallest model size ($3.9k$ parameters) among all of the compared point cloud upsampling methods, which is less than $60\%$ of Grad-PU ($6.7k$ parameters) holding the title before this study. Fig. \ref{fig:memory} shows how our memory efficiency and performances stand compared to the compared methods. Our method is consistently placed at the bottom left, indicating that it achieves SOTA performance without using up a substantial computational overhead. Considering the case of self-supervised learning, choosing our method can significantly reduce the computation requirements as SAPCU \cite{sapcu} requires $81.2k$ parameters to run its encoding and decoding processes compared to 38,916 parameters in our method. We experiment with our supervised model trained with a reduced batch size to adapt to a small GPU for fair comparisons of the training computation costs between different methods. Our method generally shows faster training and inference speeds, except for the inference speed of Grad-PU \cite{gradpu} on the urban (KITTI-360) dataset (Table \ref{table:computation}). Our method also shows the smallest gap between inference speeds in urban and highway environments. Processing on large-scale point clouds is performed on GPUs with memory sizes of $10 \sim11$ gigabytes.

\pgfplotsset{width=4cm,compat=1.18}
\begin{figure}[!htbp]
    \centering
    \subfloat[CD]{
        \begin{tikzpicture}[background rectangle/.style={draw=red, thick}]
          \begin{axis}[
    ymode=log,
    ylabel=Model Size,
    xlabel near ticks,
    ylabel near ticks,
    label style={inner sep=0pt, font=\fontsize{6}{6}\selectfont},
    y tick label style={font=\fontsize{6}{6}\selectfont, inner sep=0pt},
    x tick label style={font=\fontsize{6}{6}\selectfont},
]
    \addplot[color=orange, only marks, mark=asterisk]
    coordinates{(0.493, 812003)}; \label{punet}
    \addplot[color=brown, only marks, mark=x]
    coordinates{(0.305, 76155)}; \label{mpu}
    \addplot[color=purple, only marks, mark=+]
    coordinates{(0.296, 541601)}; \label{pugan}
    \addplot[color=yellow, only marks, mark=triangle]
    coordinates{(0.291, 75971)}; \label{pugcn}
    \addplot[color=blue, only marks, mark=o]
    coordinates{(0.465, 8718348)}; \label{sapcu}
    \addplot[color=green, only marks, mark=square]
    coordinates{(0.268, 67073)}; \label{gradpu}
    \addplot[color=red, only marks, mark=diamond]
    coordinates{(0.260, 38916)}; \label{puray}

  \end{axis}
        \end{tikzpicture}
    }
    \subfloat[HD]{
        \begin{tikzpicture}[background rectangle/.style={draw=red, thick}]
        \begin{axis}[
    ymode=log,
    yticklabels={,,},
    xlabel near ticks,
    ylabel near ticks,
    label style={inner sep=0pt, font=\fontsize{6}{6}\selectfont},
    y tick label style={font=\fontsize{6}{6}\selectfont, inner sep=0pt},
    x tick label style={font=\fontsize{6}{6}\selectfont},
]
    \addplot[color=orange, only marks, mark=asterisk]
    coordinates{(4.508, 812003)}; \label{punet}
    \addplot[color=brown, only marks, mark=x]
    coordinates{(4.463, 76155)}; \label{mpu}
    \addplot[color=purple, only marks, mark=+]
    coordinates{(5.722, 541601)}; \label{pugan}
    \addplot[color=yellow, only marks, mark=triangle]
    coordinates{(2.986, 75971)}; \label{pugcn}
    \addplot[color=blue, only marks, mark=o]
    coordinates{(10.572, 8718348)}; \label{sapcu}
    \addplot[color=green, only marks, mark=square]
    coordinates{(2.601, 67073)}; \label{gradpu}
    \addplot[color=red, only marks, mark=diamond]
    coordinates{(2.736, 38916)}; \label{puray}
  \end{axis}
        \end{tikzpicture}
    }
    \subfloat[P2F]{
        \begin{tikzpicture}[background rectangle/.style={draw=red, thick}]
          \begin{axis}[
    ymode=log,
    yticklabels={,,},
    xlabel near ticks,
    ylabel near ticks,
    label style={inner sep=0pt, font=\fontsize{6}{6}\selectfont},
    y tick label style={font=\fontsize{6}{6}\selectfont, inner sep=0pt},
    x tick label style={font=\fontsize{6}{6}\selectfont},
  ]
    \addplot[color=orange, only marks, mark=asterisk]
    coordinates{(4.315, 812003)}; \label{punet}
    \addplot[color=brown, only marks, mark=x]
    coordinates{(2.882, 76155)}; \label{mpu}
    \addplot[color=purple, only marks, mark=+]
    coordinates{(2.812, 541601)}; \label{pugan}
    \addplot[color=yellow, only marks, mark=triangle]
    coordinates{(2.471, 75971)}; \label{pugcn}
    \addplot[color=blue, only marks, mark=o]
    coordinates{(3.421, 8718348)}; \label{sapcu}
    \addplot[color=green, only marks, mark=square]
    coordinates{(1.990, 67073)}; \label{gradpu}
    \addplot[color=red, only marks, mark=diamond]
    coordinates{(1.707, 38916)}; \label{puray}
  \end{axis}
        \end{tikzpicture}
    }
    \newline
    \includegraphics[width=\columnwidth]{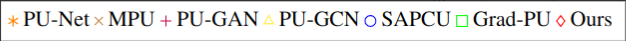}
    \caption{Number of parameters versus upsampling performances of PU-Ray and compared models.}
    \label{fig:memory}
\end{figure}
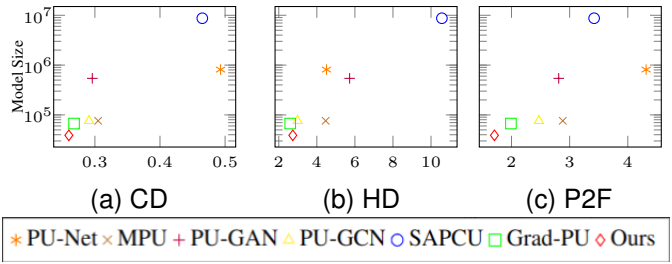

\subsection{Ablation studies}
\pgfplotsset{width=4.9cm,compat=1.18}
\begin{figure}[!tbp]
    \centering
    \subfloat{
        \begin{tikzpicture}[background rectangle/.style={draw=red, thick}]
        \begin{axis}[
    x tick label style={},
    xlabel={Epoch},
    ylabel={MAE},
    scaled x ticks=false,
    xmin=0, xmax=100,
    ymin=0.02, ymax=0.06,
    xtick={20, 40, 60, 80, 100},
    ytick={0.02, 0.03, 0.04, 0.05, 0.06},
    legend pos=north east,
    legend style={font=\fontsize{6}{6}\selectfont},
    ymajorgrids=true,
    grid style=dashed,
]

\addplot[
    color=blue,
    ]
    coordinates {
        (0, 0.057324044863737496)
        (1, 0.04404288355927534)
        (2, 0.0420645197254984)
        (3, 0.03884490817614765)
        (4, 0.037696910042866134)
        (5, 0.0367769941852185)
        (6, 0.03473549250521287)
        (7, 0.034014992569220026)
        (8, 0.033459795854326874)
        (9, 0.03575097084304949)
        (10, 0.03279891241461382)
        (11, 0.0323335314308362)
        (12, 0.03220642485947146)
        (13, 0.03253933963372489)
        (14, 0.031812431862032794)
        (15, 0.03237160785492648)
        (16, 0.03208704826113678)
        (17, 0.03262831427462846)
        (18, 0.03120749746458705)
        (19, 0.031009055678573847)
        (20, 0.031103707292458177)
        (21, 0.031108784378419158)
        (22, 0.03094419315599305)
        (23, 0.03090746542024997)
        (24, 0.03159965898319984)
        (25, 0.031307681256097335)
        (26, 0.03026800974798755)
        (27, 0.03102120126893073)
        (28, 0.030595311969403807)
        (29, 0.030822091156068883)
        (30, 0.029702350761510156)
        (31, 0.029169804622673188)
        (32, 0.030387275459382974)
        (33, 0.029121198535368428)
        (34, 0.029029337508331934)
        (35, 0.02908866195988977)
        (36, 0.02845775968506986)
        (37, 0.028846438012235694)
        (38, 0.028865409343689623)
        (39, 0.028480179036798524)
        (40, 0.028849288094579286)
        (41, 0.029271062505884307)
        (42, 0.028448140399768666)
        (43, 0.02789252027928471)
        (44, 0.02892261913222933)
        (45, 0.027958895179508718)
        (46, 0.028285429943580947)
        (47, 0.02858726578924064)
        (48, 0.028544006647641805)
        (49, 0.027810321501958727)
        (50, 0.02754806993548181)
        (51, 0.027535775532604997)
        (52, 0.027847790507411065)
        (53, 0.02727289882060785)
        (54, 0.02748288386005383)
        (55, 0.027682554699461855)
        (56, 0.028079879005843602)
        (57, 0.02736571161941228)
        (58, 0.02725907252705012)
        (59, 0.027281089765710788)
        (60, 0.027169352315068358)
        (61, 0.027262517199003113)
        (62, 0.027066958631560577)
        (63, 0.02694337122855823)
        (64, 0.02703733115295966)
        (65, 0.02702956067111065)
        (66, 0.026848068550773103)
        (67, 0.027092883257531807)
        (68, 0.02674392002172229)
        (69, 0.026599833322090132)
        (70, 0.027217798108014276)
        (71, 0.02688167510042555)
        (72, 0.026656629889306012)
        (73, 0.02652222860039403)
        (74, 0.026665213931131426)
        (75, 0.026527117155684055)
        (76, 0.02675982021161657)
        (77, 0.026639799390760423)
        (78, 0.026619769805939796)
        (79, 0.02665304372829812)
        (80, 0.026997450520675998)
        (81, 0.026622737508602976)
        (82, 0.026692266108219665)
        (83, 0.026439047780304255)
        (84, 0.026402854916191362)
        (85, 0.02658867876747434)
        (86, 0.02640466149657007)
        (87, 0.026542541834156937)
        (88, 0.026534541956073278)
        (89, 0.026151979383441517)
        (90, 0.026202360055485343)
        (91, 0.02625938695893471)
        (92, 0.02627879088324325)
        (93, 0.02619913688893951)
        (94, 0.026193090291067398)
        (95, 0.026160814448880593)
        (96, 0.02596863955695181)
        (97, 0.026221017478200173)
        (98, 0.02604623196182259)
        (99, 0.027018752239667078)
    };
    \legend{\# steps = 0}

\addplot[
    color=red,
    ]
    coordinates {
        (0, 0.05901111873048548)
        (1, 0.051492293816571354)
        (2, 0.04598807309953299)
        (3, 0.040739987177211515)
        (4, 0.03994522093619041)
        (5, 0.035253851026039874)
        (6, 0.03317281733472948)
        (7, 0.03219793714894174)
        (8, 0.034712284165843627)
        (9, 0.03047488630016931)
        (10, 0.029421840069676592)
        (11, 0.03100527106408397)
        (12, 0.028861227045048145)
        (13, 0.02776303879269909)
        (14, 0.027949376022897994)
        (15, 0.028101674769929516)
        (16, 0.027067011220963534)
        (17, 0.02644475368007991)
        (18, 0.02636607918813227)
        (19, 0.030043547962098873)
        (20, 0.025803790267238627)
        (21, 0.025603492903579588)
        (22, 0.02597830522628998)
        (23, 0.025323483438242325)
        (24, 0.025020848519474857)
        (25, 0.02508379248514658)
        (26, 0.025109702773117505)
        (27, 0.024715794590071758)
        (28, 0.024671919403813896)
        (29, 0.024996637711925344)
        (30, 0.024960790956602554)
        (31, 0.02437353310430982)
        (32, 0.025045518672904672)
        (33, 0.02423827885866233)
        (34, 0.02422049164541948)
        (35, 0.024520502495147107)
        (36, 0.024032339298847738)
        (37, 0.023979999608689875)
        (38, 0.02363186711596232)
        (39, 0.024305247292480125)
        (40, 0.023773033458972657)
        (41, 0.02370199017220989)
        (42, 0.023621041551532433)
        (43, 0.023458571034858267)
        (44, 0.023489931653762234)
        (45, 0.023315394872126177)
        (46, 0.023138944080972754)
        (47, 0.023283230736067726)
        (48, 0.023485876266067886)
        (49, 0.023105697847244894)
        (50, 0.023090400230721765)
        (51, 0.02298284905467895)
        (52, 0.02312869119864825)
        (53, 0.02309059203629769)
        (54, 0.022679458817005182)
        (55, 0.02296956240081466)
        (56, 0.022774419620833757)
        (57, 0.02308511189364872)
        (58, 0.022694767445797827)
        (59, 0.022716256477396973)
        (60, 0.022664412115681536)
        (61, 0.022992970427378214)
        (62, 0.022499772423103033)
        (63, 0.022464255987570583)
        (64, 0.022738970384679317)
        (65, 0.022892324575302443)
        (66, 0.022299832013693767)
        (67, 0.022705890137425903)
        (68, 0.022155591040462993)
        (69, 0.022772265403053084)
        (70, 0.022435109736443245)
        (71, 0.02262097403994379)
        (72, 0.022540640826236313)
        (73, 0.02209693645314063)
        (74, 0.02220475930292351)
        (75, 0.02227775221711074)
        (76, 0.02223961472323011)
        (77, 0.022472957039015896)
        (78, 0.022174016349731332)
        (79, 0.022171310776386986)
        (80, 0.0224919511532207)
        (81, 0.02196822699958543)
        (82, 0.02201588029671186)
        (83, 0.021978799264215535)
        (84, 0.022288839445570006)
        (85, 0.022051404721450504)
        (86, 0.021987728708661284)
        (87, 0.0231412555896924)
        (88, 0.021985869103532812)
        (89, 0.02217217208366031)
        (90, 0.022152919437582144)
        (91, 0.021794801814194076)
        (92, 0.02206019018107989)
        (93, 0.021637312125894523)
        (94, 0.022233984741469644)
        (95, 0.02197745041034713)
        (96, 0.02201818966267347)
        (97, 0.022018308134574522)
        (98, 0.021682047183712878)
        (99, 0.02173565432636089)
    };
    
\legend{w/o Ray Marching, w/ Ray Marching}
\end{axis}
        \end{tikzpicture}
    }
    \subfloat{
        \begin{tikzpicture}[background rectangle/.style={draw=red, thick}]
        \begin{axis}[
    x tick label style={},
    xlabel={Epoch},
    ylabel={RMSE},
    scaled x ticks=false,
    xmin=0, xmax=100,
    ymin=0.03, ymax=0.08,
    xtick={20, 40, 60, 80, 100},
    ytick={0.03, 0.04, 0.05, 0.06, 0.07, 0.08},
    legend pos=north east,
    legend style={font=\fontsize{6}{6}\selectfont},
    ymajorgrids=true,
    grid style=dashed,
]

\addplot[
    color=blue,
    ]
    coordinates {
        (0, 0.0771398349750359)
        (1, 0.06163288020902387)
        (2, 0.05843924264484825)
        (3, 0.05524647408728346)
        (4, 0.0540324207113492)
        (5, 0.05250942927739362)
        (6, 0.050252091186156)
        (7, 0.04925232557070395)
        (8, 0.0486796572241382)
        (9, 0.05022167397021898)
        (10, 0.04775394403975786)
        (11, 0.04697940462697065)
        (12, 0.046811917403067754)
        (13, 0.046906369212965246)
        (14, 0.046163827092501615)
        (15, 0.04655299872502654)
        (16, 0.046424806045023845)
        (17, 0.04624973613974464)
        (18, 0.04479427832636368)
        (19, 0.04468021032067752)
        (20, 0.04469657398662335)
        (21, 0.044991510721521154)
        (22, 0.04447178217861118)
        (23, 0.043972063448682056)
        (24, 0.04513757147889401)
        (25, 0.044658318929832845)
        (26, 0.04341553832493844)
        (27, 0.04401887438996867)
        (28, 0.043559249105739724)
        (29, 0.044083772632866815)
        (30, 0.042407900763469515)
        (31, 0.04206818902387166)
        (32, 0.0435336839401628)
        (33, 0.04189282775014402)
        (34, 0.041824739710519705)
        (35, 0.042081146407740876)
        (36, 0.041120931556182266)
        (37, 0.04174763957297776)
        (38, 0.041352686153008904)
        (39, 0.04113034502286143)
        (40, 0.0418073176651562)
        (41, 0.04204144457157974)
        (42, 0.04092767389647898)
        (43, 0.040422289084996746)
        (44, 0.04127686806872374)
        (45, 0.04052939796995808)
        (46, 0.04061353696754933)
        (47, 0.04118110099213695)
        (48, 0.04094752147593146)
        (49, 0.040391188674226614)
        (50, 0.03997492420122085)
        (51, 0.039886512595287756)
        (52, 0.04049452587370058)
        (53, 0.03959555555257383)
        (54, 0.03991863468532141)
        (55, 0.039905450859580686)
        (56, 0.04029032354417843)
        (57, 0.03955315097278373)
        (58, 0.039784681076529126)
        (59, 0.039473697329514064)
        (60, 0.03947899377316098)
        (61, 0.03947779262239267)
        (62, 0.03930121414449136)
        (63, 0.03906432638875222)
        (64, 0.03929917881460499)
        (65, 0.03934894228748886)
        (66, 0.038815657656878916)
        (67, 0.039167915468503636)
        (68, 0.03905939986784062)
        (69, 0.038830383073821556)
        (70, 0.039428933699148454)
        (71, 0.03897388386862357)
        (72, 0.038723597967789014)
        (73, 0.038828902383888045)
        (74, 0.038853930926136766)
        (75, 0.03857530344727129)
        (76, 0.03882421456151247)
        (77, 0.03885995134359205)
        (78, 0.03879517778001986)
        (79, 0.03889625701146117)
        (80, 0.03909523922285536)
        (81, 0.03875252454693643)
        (82, 0.038699931874787805)
        (83, 0.038624435605903853)
        (84, 0.038482623560094846)
        (85, 0.038782994928525105)
        (86, 0.03838812855079894)
        (87, 0.03858169217791086)
        (88, 0.038554109304761316)
        (89, 0.0381690754532321)
        (90, 0.03836645944127785)
        (91, 0.03814466603831914)
        (92, 0.03834328527066376)
        (93, 0.03811166086376532)
        (94, 0.03808737472795603)
        (95, 0.03818310003207224)
        (96, 0.03803249602206571)
        (97, 0.03813318159928824)
        (98, 0.03792967600526831)
        (99, 0.03890737606230798)
    };
    \legend{\# steps = 0}

\addplot[
    color=red,
    ]
    coordinates {
        (0, 0.0792082299639875)
        (1, 0.0710185106149289)
        (2, 0.063294836324500)
        (3, 0.057561956357018)
        (4, 0.055668986040140)
        (5, 0.0503485587691793)
        (6, 0.0472066466857355)
        (7, 0.0455633035628676)
        (8, 0.05002095993979717)
        (9, 0.0435419270410280)
        (10, 0.04224939284249666)
        (11, 0.0433395081968283)
        (12, 0.04104511004590192)
        (13, 0.0400799242490046)
        (14, 0.0406852252312845)
        (15, 0.0403285912855980)
        (16, 0.0392828412513118)
        (17, 0.0387050102545533)
        (18, 0.0384648761825530)
        (19, 0.0424730341219122)
        (20, 0.0379686449388587)
        (21, 0.0377108177212520)
        (22, 0.03791059816958474)
        (23, 0.0372740760191419)
        (24, 0.037129934684131)
        (25, 0.03717637585139136)
        (26, 0.03706373911259)
        (27, 0.03680516181700694)
        (28, 0.0367742611132039)
        (29, 0.03704586037735719)
        (30, 0.0368723730844202)
        (31, 0.0363571392645694)
        (32, 0.0369869314509051)
        (33, 0.0362904328479273)
        (34, 0.0361600334060815)
        (35, 0.0367682737322444)
        (36, 0.0360754223590038)
        (37, 0.03588013571772529)
        (38, 0.0357383390054911)
        (39, 0.0362489667241671)
        (40, 0.03577967711628251)
        (41, 0.03557028536895049)
        (42, 0.0353790104285237)
        (43, 0.0352807022984418)
        (44, 0.03544387146886468)
        (45, 0.0352224294963498)
        (46, 0.03507823155363952)
        (47, 0.03519823348545010)
        (48, 0.035669265857970)
        (49, 0.0349342766026941)
        (50, 0.0349185199359225)
        (51, 0.0348395206912826)
        (52, 0.03510470961574977)
        (53, 0.0352956235782130)
        (54, 0.0346671492654428)
        (55, 0.0347695251380317)
        (56, 0.03462753267983751)
        (57, 0.0348724360783120)
        (58, 0.0347131424828757)
        (59, 0.0345701214032025)
        (60, 0.03455972289375043)
        (61, 0.0346259699100029)
        (62, 0.0343458674536304)
        (63, 0.0343484505321370)
        (64, 0.03435519266675076)
        (65, 0.0348659309821017)
        (66, 0.034106389724477)
        (67, 0.0344802091737947)
        (68, 0.0340794376189276)
        (69, 0.0344987974417946)
        (70, 0.0343914023497604)
        (71, 0.0343679137728646)
        (72, 0.03456579696184281)
        (73, 0.0339595007406691)
        (74, 0.0339992763154838)
        (75, 0.03390220948704794)
        (76, 0.03409781024403977)
        (77, 0.0343311321252171)
        (78, 0.0341147973159088)
        (79, 0.03396359169861242)
        (80, 0.0341550577140939)
        (81, 0.0337416617770471)
        (82, 0.03378841941507150)
        (83, 0.03389183485236777)
        (84, 0.034029133598556)
        (85, 0.03389612476901946)
        (86, 0.03372588282808966)
        (87, 0.03507961429447334)
        (88, 0.03375653269542934)
        (89, 0.0338330951880124)
        (90, 0.034025776932735)
        (91, 0.0336575073293215)
        (92, 0.0336853552740085)
        (93, 0.0333824134342673)
        (94, 0.0340008070682541)
        (95, 0.0335351521286335)
        (96, 0.0337906145512183)
        (97, 0.03374732908023094)
        (98, 0.0335638968558909)
        (99, 0.03344130656272549)
    };

\legend{w/o Ray Marching, w/ Ray Marching}
\end{axis}
        \end{tikzpicture}
    }
    \caption{Learning curve graphs of the ray marching module on validation depth MAE and RMSE with and without the sphere tracing algorithm.}
    \label{fig:validation}
\end{figure}
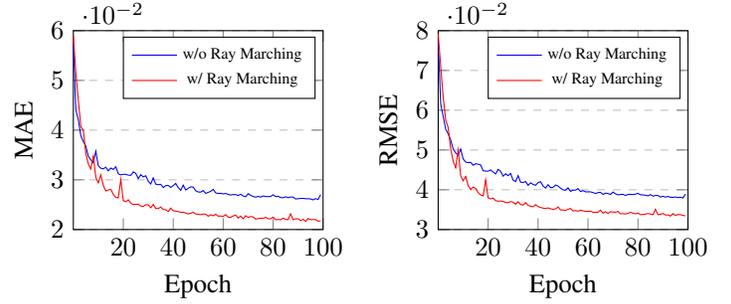

In ablation studies, we justify the choice of designs for our method. Fig. \ref{fig:validation} shows the validation mean absolute errors (MAE) and root mean squared errors (RMSE) of the depth predictions during training of the ray marching module with and without the spherical tracing algorithm \cite{sphere_tracing}. The consistent results of MAE and RMSE suggest that the ray marching positively affects representation learning of the neural implicit surface through multiple iterations of UDF definition.

\pgfplotsset{width=6cm,compat=1.9}
\begin{figure}[!htbp]
\center
\begin{tikzpicture}
\let\center\empty
\let\endcenter\relax
\centering
\begin{axis}[
    xmode=log,
    log basis x={2},
    x tick label style={},
    xlabel={Number of training queries},
    ylabel={MAE},
    scaled x ticks=false,
    xmin=2048, xmax=16384,
    ymin=0.028, ymax=0.042,
    xtick={2048,4096,8192,16384},
    ytick={0.028,0.030,0.032,0.034,0.036, 0.038, 0.040, 0.042},
    legend pos=outer north east,
    ymajorgrids=true,
    grid style=dashed,
]

\addplot[
    color=blue,
    mark=square,
    ]
    coordinates {
    (2048, 0.04104866453)(4096, 0.03755705383)(8192, 0.03498130302)(16384, 0.03301250174)
    };
    \legend{\# steps = 0}

\addplot[
    color=red,
    mark=square,
    ]
    coordinates {
    (2048, 0.04150525486)(4096, 0.0381133939)(8192, 0.03501970206)(16384, 0.03209722998)
    };

\addplot[
    color=yellow,
    mark=square,
    ]
    coordinates {
    (2048, 0.03951351786)(4096, 0.0355055892)(8192, 0.03234689879)(16384, 0.02948867205)
    };

\addplot[
    color=green,
    mark=square,
    ]
    coordinates {
    (2048,0.03869759027)(4096, 0.03507186504)(8192, 0.031394528)(16384, 0.02844174556)
    };

\addplot[
    color=orange,
    mark=square,
    ]
    coordinates {
    (2048, 0.03868568853)(4096, 0.03532179026)(8192, 0.03264205374)(16384, 0.02937651682)
    };
    
\legend{$M$ = 0, $M$ = 2, $M$ = 4, $M$ = 6, $M$ = 8}
\end{axis}
\end{tikzpicture}
\caption{The ablation study shows the effect of the number of ray marching steps on depth prediction performance. The depths are normalized.}
\label{fig:ablation_1}
\end{figure}
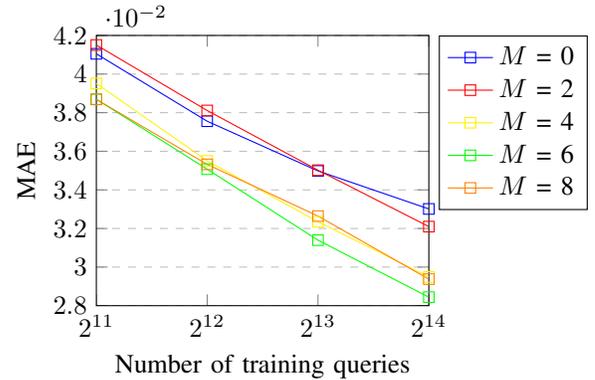

The effect of the number of ray marching steps, $M$, on the depth prediction performance is summarized in Fig. \ref{fig:ablation_1} with $M \in \{0, 2, 4, 6, 8\}$. The mean absolute error (MAE) of the depth prediction is recorded for each model trained with randomly selected query sets with sizes of 2,048, 4,096, 8,192 and 16,384. The numbers are averaged from the results of 4 trials with different random seeds for stability. As per the sphere tracing algorithm \cite{sphere_tracing}, the general trend shows that more marching steps result in better performance until six iterations throughout different training set sizes. However, the performance gain is minimal to negative from 6 to 8 marching steps, leading to our method's design choice, which is $M=6$.

\begin{table}[!htbp]
\caption{Ablation study on the loss functions. Depth MAEs are reported. The best three results are coloured red ($1^{st}$), orange ($2^{nd}$) and green ($3^{rd}$).}
\label{table:ablation_2}
\let\center\empty
\let\endcenter\relax
\centering
\begin{tabular}{lccc}
\hline
                                            & \multicolumn{3}{c}{\# of Inference Macrhing Steps} \\ 
\multicolumn{1}{c}{Loss Functions}          & 4 & 6 & 8 \\ \hline
w/ $L_{MSE}$ \& $L_{RMSE}$ only             & 0.18241 & 0.02101 & 0.08246 \\
w/o $L_{\epsilon}$                          & 0.02705 & 0.02153 & 0.02345  \\
w/o $L_{tan}$                               & \textcolor{orange}{0.02607} & \textcolor{green}{0.02005} & \textcolor{green}{0.02114}  \\
w/o $L_{ms}$                                & \textcolor{green}{0.02704} & \textcolor{orange}{0.01970} & \textcolor{orange}{0.02049} \\
Ours                                        & \textbf{\textcolor{red}{0.02526}} & \textbf{\textcolor{red}{0.01966}} & \textbf{\textcolor{red}{0.02025}} \\ \hline
\end{tabular}

\end{table}

\begin{table}[!htbp]
\caption{Ablation study on the query generation method.}
\label{table:ablation_3}
\let\center\empty
\let\endcenter\relax
\centering
\begin{tabular}{lccc}
\hline
                                            & CD $\downarrow$ & HD $\downarrow$ & P2F $\downarrow$ \\ 
\multicolumn{1}{c}{Query Point Generation Method} & $10^{-3}$ & $10^{-3}$ & $10^{-3}$ \\ \hline
Simple mid-point                            & 0.271 & 2.679 & 1.662 \\
w/ hex-neighbourhood constraint                       & 0.260 & 2.736 & 1.707 \\ \hline
\end{tabular}

\end{table}

Ablated loss functions show how each term plays a role in the depth prediction (Table \ref{table:ablation_2}). Theoretically, the depth prediction should be consistent after a few iterations following the sphere tracing algorithm \cite{sphere_tracing}. Also, a better quality UDF of the neural implicit surface is assumed to yield a precise depth prediction regardless of the number of inference marching steps. Therefore, a loss function should be rejected if the resulting model does not show such behaviour. Our novel $\epsilon$ estimation shows noticeable precision improvement to the original algorithm. Although removing one of $L_{tan}$ and $L_{ms}$ may not show significant degradation of the final performance, they are essential for following the original algorithm and generating implicit surface representation, as the results do not show consistency in different marching steps. The hexagonal neighbourhood constraint for query generation affects the performance gain in the Chamfer Distance with the trade-offs in the Hausdorff and P2F distances Table. \ref{table:ablation_3}.

\section{Limitations}
Although PU-Ray takes advantage of the freedom of ROIs definition and our mid-point algorithm shows successful upsampling results, the quality of upsampling results relies heavily on the query ray and patch generation. There are potential occasions where it is unsuitable when the surface is too complex. Further modifications may be required to overcome this problem.

While $k$-NN operations are widely accepted by many other upsampling methods \cite{pu_net, mpu, pu_gcn, sspu, spu_net, sapcu, pugac, pu_transformer, gradpu}, it is only an approximation of the local surface. It is inadequate for query ray and patch generation when the surface is too complex. Also, the process is computationally expensive and not ideal in time-critical applications. Future work should focus on generating more precise and computationally efficient local patches. 

\section{Conclusion}
This paper addresses the domain dependency problem of existing end-to-end point cloud upsampling methods by proposing a new ray-based upsampling method, PU-Ray. The proposed method demonstrates that the sphere tracing algorithm \cite{sphere_tracing} on a neural implicit surface yields a precise query ray depth prediction. An arbitrary number of query rays are generated using a novel rule-based scheme, which resolves the non-uniform distribution using existing models trained with nearest-neighbor-based reconstruction loss functions. Our method also takes advantage of the accurate ground truth for training query rays in both supervised and self-supervised learning so that it is applicable to real-world ITS scenarios. Our method achieves SOTA quantitative results on synthetic datasets and noisy inputs while having the smallest number of parameters. Moreover, our comprehensive experiments on novel datasets for 3D reconstruction suggest the method's robustness in urban and highway environments compared to other upsampling methods. The empirical evidence suggests that upsampling should be considered over range-image super-resolution for 3D reconstruction. Future research may benefit from our findings of ray marching on point clouds for precise depth predictions on real-scanned surfaces. Although this research demonstrates the potential significance of point cloud upsampling to the ITS domain, all the methods, including PU-Ray, have inference speeds of lab experimentations with limited computation resources, and application on real-scanned data remains challenging. Future investigations to accelerate the processes to satisfy industrial standards are suggested.

\bibliographystyle{IEEEtran}
\bibliography{puray}

\begin{IEEEbiography}[{\includegraphics[width=1in,height=1.25in,clip,keepaspectratio]{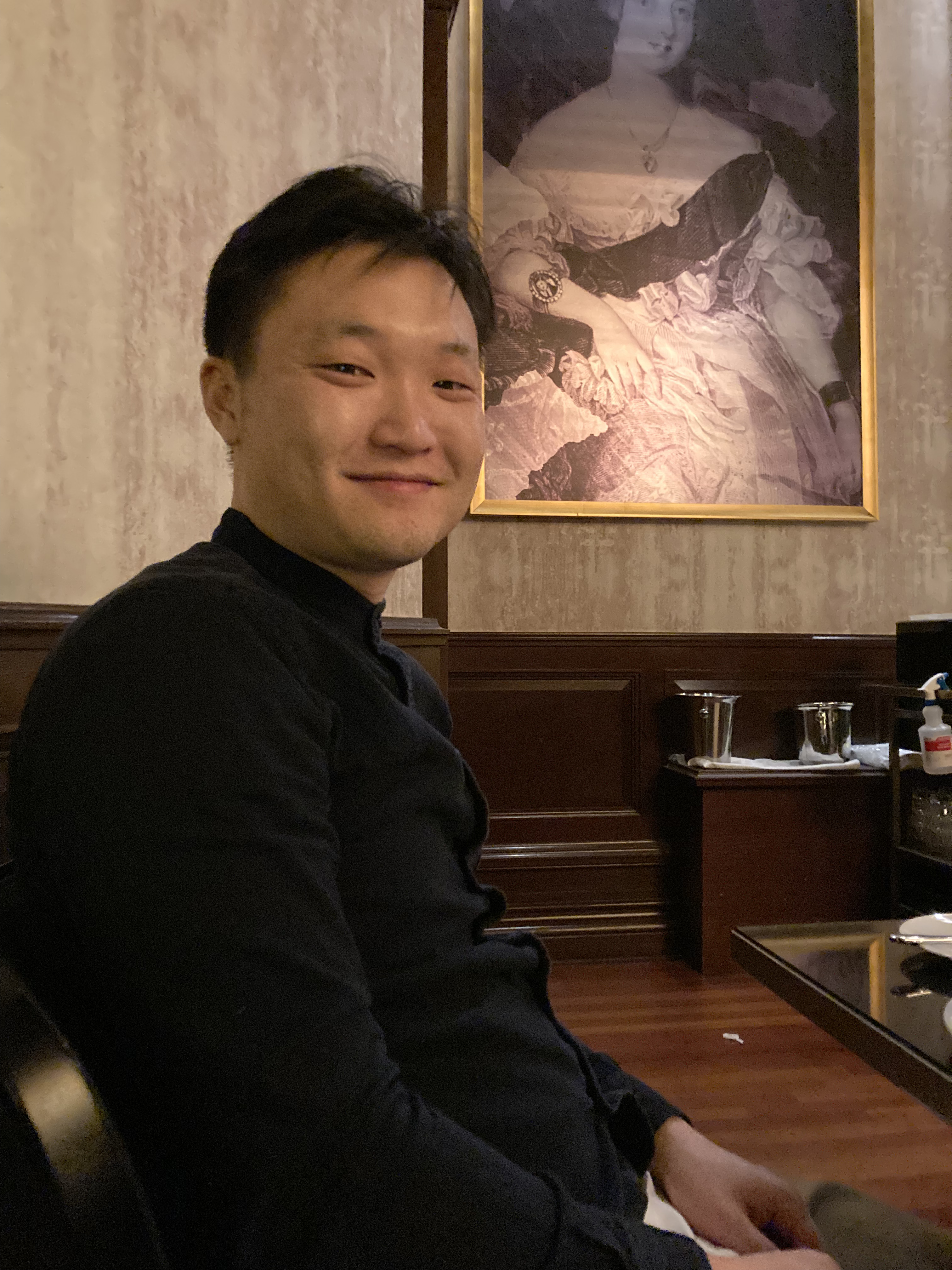}}]{Sangwon Lim}
received a B.Sc. degree in Computer Science and Geography (Geomatics) from the University of Victoria, Victoria, BC, Canada, in 2022. He is currently pursuing an M.Sc. degree in Computing Science at the University of Alberta, Edmonton, AB, Canada. His research areas are computer vision and graphics for remote sensing data enhancement. Projects at the Centre for Smart Transportation (CST) investigate Civil Engineering infrastructure adaptability to ITS.
\end{IEEEbiography}

\begin{IEEEbiography}[{\includegraphics[width=1in,height=1.25in, clip,keepaspectratio]{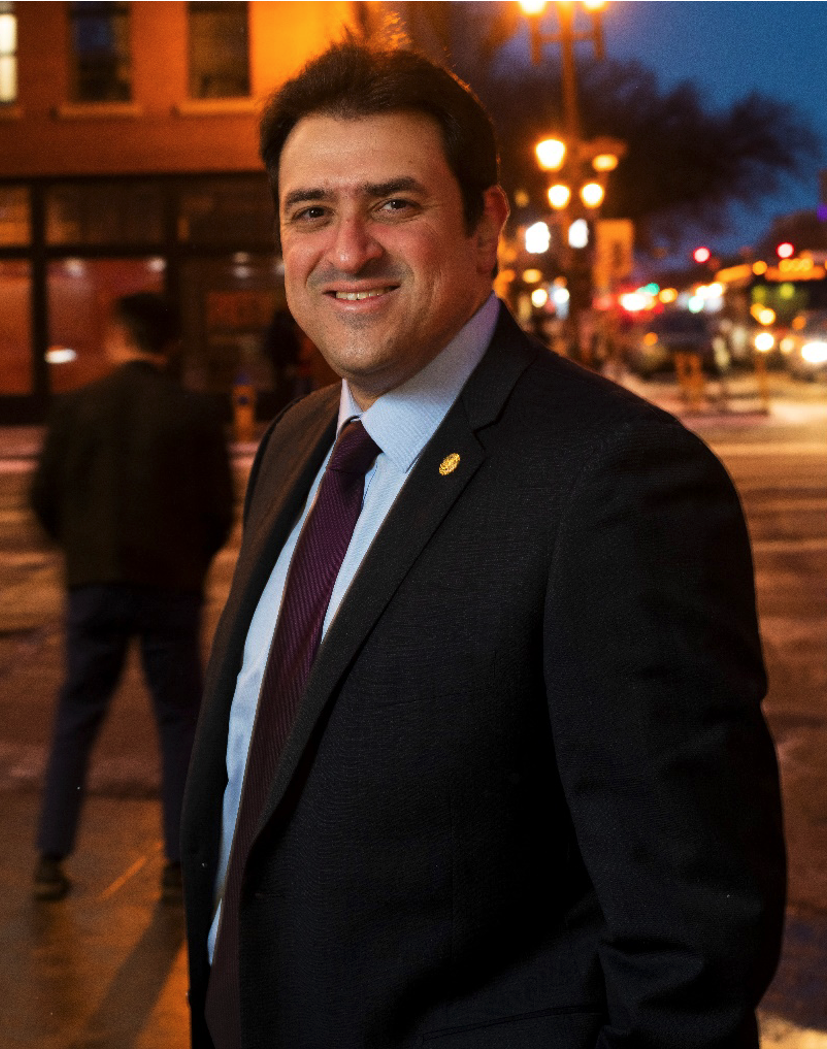}}]{Karim El-Basyouny} is a professor and inaugural City of Edmonton’s Research Chair in Urban Traffic Safety at the University of Alberta (UofA). He is a co-founder and steering committee member for the Centre of Smart Transportation and serves as the Associate Dean for Research in the Faculty of Engineering. He joined the UofA in July 2011 after completing his MASc and Ph.D. in Transportation Engineering from the University of British Columbia. His research program focuses on advanced road safety management. This includes a comprehensive body of work on developing integrated speed management strategies and infrastructure digitization for Connected and Automated Vehicles. He is an active member of multiple (inter)national safety committees and serves on the editorial boards of several prominent journals. Dr. El-Basyouny has won several notable research awards throughout his career. His most recent is the Canadian ITS Diversity, Equity, \& Inclusion Award, the UofA’s Faculty of Engineering Research Award, and the Canadian ITS R\&D/Innovation Award.
\end{IEEEbiography}

\begin{IEEEbiography}[{\includegraphics[width=1in,height=1.25in, clip,keepaspectratio]{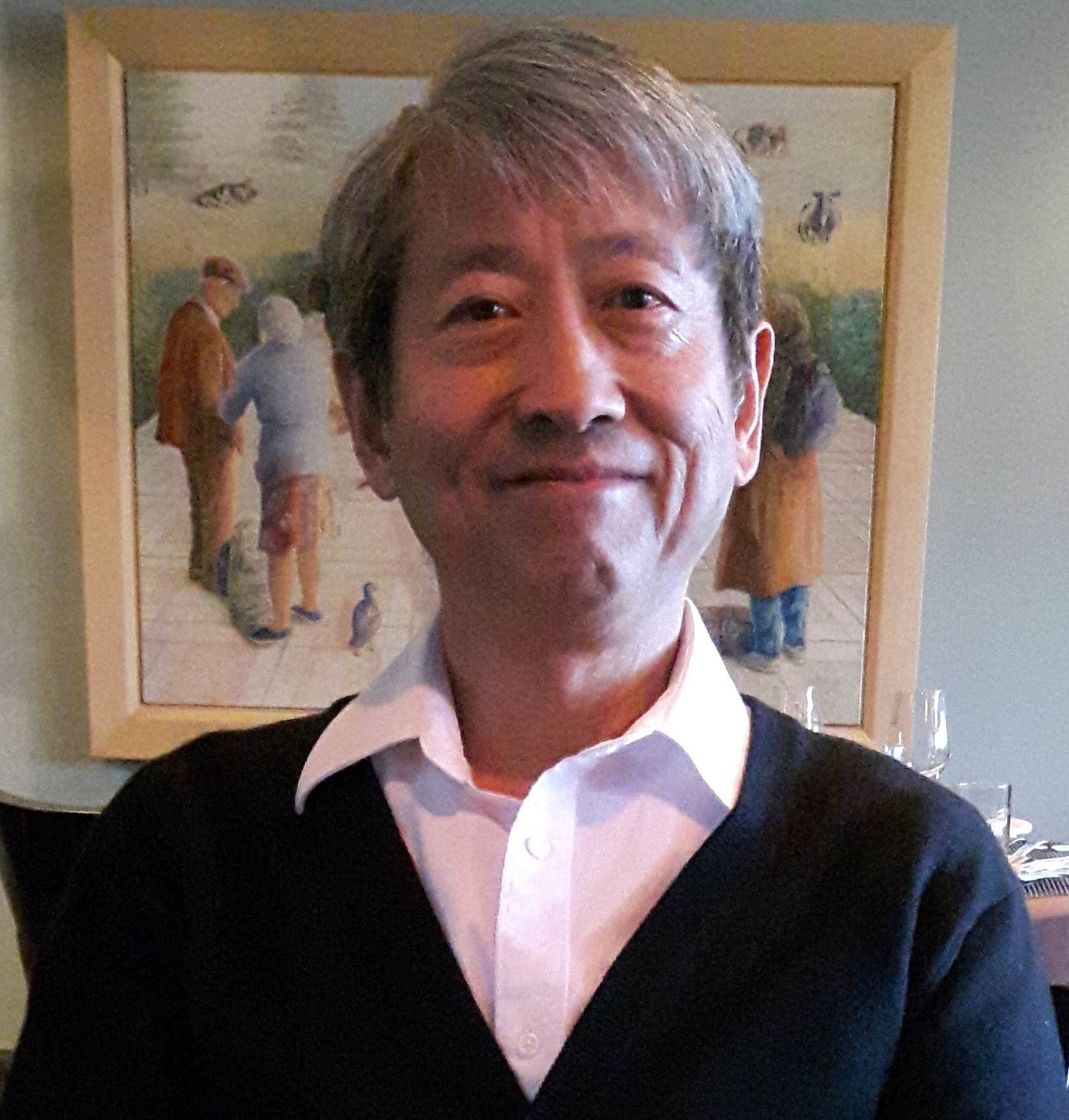}}]{Yee Hong Yang}'s research interests cover a wide range of topics in computer graphics and computer vision. In computer graphics, his interests include fluid and character animation, environment matting, hardware-accelerated graphics, motion editing, physics-based modeling, texture analysis and synthesis, and static and dynamic image-based modeling and rendering. His research in computer vision encompasses light source estimation, motion estimation, human motion analysis, camera calibration, 3D reconstruction, stereo and multiview stereo, underwater imaging, and medical imaging, particularly in developing biomarkers for ALS. He is a senior member of the IEEE and serves on the Editorial Board of the journal Pattern Recognition. He has published over 160 technical papers in international journals and conference proceedings, co-edited one book, and served as a guest editor of an international journal. In addition, he has served as a reviewer for many international journals and as a committee member for many conferences, review panels, and government agencies.
\end{IEEEbiography}

\end{document}


\title{Supplementary Material of "PU-Ray: Point Cloud Upsampling via Ray Marching on Implicit Surface"}

\author{
    Sangwon Lim,
    \thanks{Sangwon Lim is with the Centre for Smart Transportation (CST) and the Department of Computing Science at the University of Alberta, Edmonton, AB, Canada (e-mail: sangwon3@ualberta.ca).}
  Karim El-Basyouny,
    \thanks{Karim El-Basyouny is with the Centre for Smart Transportation (CST) and the Department of Civil and Environmental Engineering, University of Alberta, Edmonton, AB, Canada T6G 1H9 (e-mail: basyouny@ualberta.ca).}
  and
  Yee Hong Yang
    \thanks{Yee Hong Yang is with the Department of Computing Science, University of Alberta, Edmonton, AB, Canada T6G 2E8 (e-mail: herberty@ualberta.ca).}
}



\renewcommand\fbox{\fcolorbox{red}{white}}
\setlength{\fboxrule}{2pt} 

\newcommand\submittedtext{%
  \footnotesize This work has been submitted to the IEEE for possible publication. Copyright may be transferred without notice, after which this version may no longer be accessible.}

\newcommand\submittednotice{%
\begin{tikzpicture}[remember picture,overlay]
\node[anchor=south,yshift=10pt] at (current page.south) {\fbox{\parbox{\dimexpr0.65\textwidth-\fboxsep-\fboxrule\relax}{\submittedtext}}};
\end{tikzpicture}%
}

\maketitle
\submittednotice


{
\section*{Justification of $L_\epsilon$}

Instead of $L_\epsilon$, one may suggest the ReLu activation, which outputs $ReLu(\epsilon^{\textbf{q}}) = max(0, \epsilon^{\textbf{q}})$, at the last layer to obtain non-negative epsilon values naturally. However, this may introduce the dying ReLu problem \cite{dyingrelu}, where the network does not update when the $\frac{\partial L}{\partial a^\textbf{q}} = 0$, where $a^\textbf{q}=ReLu(\epsilon^{\textbf{q}})$. For example, given that,
\begin{align*} \label{eq:supp_1} \tag{Supp. 1}
&L_{MAE} = \frac{\sum^{Q}_{\textbf{q}} |\hat{t}^{\textbf{q}} - t^{\textbf{q}}|}{|Q|}, \\ 
&L_{RMSE} = \sqrt{\frac{\sum^{Q}_{\textbf{q}} (\hat{t}^{\textbf{q}}-t^{\textbf{q}})^2}{|Q|}},
\end{align*}
we have,
\begin{align*} \label{eq:supp_2} \tag{Supp. 2}
\frac{\partial{L_{MAE}}}{\partial a^{\textbf{q}'}} &= \frac{1}{|Q|} \cdot \sum^{Q}_\textbf{q} \frac{\partial}{\partial a^{\textbf{q}'}} |\hat{t}^{\textbf{q}} - t^{\textbf{q}}|, \\
&= \frac{1}{|Q|} \cdot \sum^{Q}_\textbf{q} \frac{\partial}{\partial a^{\textbf{q}'}} |\Tilde{t}^{\textbf{q}} + a^{\textbf{q}} - t^{\textbf{q}}|, \\
&= \frac{1}{|Q|} \cdot \sum^{Q}_\textbf{q} \frac{(\Tilde{t}^{\textbf{q}} + a^{\textbf{q}} - t^{\textbf{q}}) \cdot \frac{\partial}{\partial a^{\textbf{q}'}} (\Tilde{t}^{\textbf{q}} + a^{\textbf{q}} - t^{\textbf{q}})}{|\Tilde{t}^{\textbf{q}} + a^{\textbf{q}} - t^{\textbf{q}}|}, \\
&= \frac{1}{|Q|} \cdot \frac{(\Tilde{t}^{\textbf{q}'} + a^{\textbf{q}'} - t^{\textbf{q}'}) \cdot a^{\textbf{q}'}}{|\Tilde{t}^{\textbf{q}'} + a^{\textbf{q}'} - t^{\textbf{q}'}|},
\end{align*}
and
\begin{align*} \label{eq:supp_3} \tag{Supp. 3}
\frac{\partial{L_{RMSE}}}{\partial a^{\textbf{q}'}} &= C \cdot \frac{1}{|Q|} \cdot \sum^{Q}_\textbf{q} \frac{\partial}{\partial a^{\textbf{q}'}} (\hat{t}^{\textbf{q}} - t^{\textbf{q}})^2, \\
&= C \cdot \frac{1}{|Q|} \cdot \sum^{Q}_\textbf{q} \frac{\partial}{\partial a^{\textbf{q}'}} (\Tilde{t}^{\textbf{q}} + a^{\textbf{q}} - t^{\textbf{q}})^2, \\
&= C \cdot \frac{1}{|Q|} \cdot \sum^{Q}_\textbf{q} 2(\Tilde{t}^{\textbf{q}} + a^{\textbf{q}} - t^{\textbf{q}}) \frac{\partial}{\partial a^{\textbf{q}'}} (\Tilde{t}^{\textbf{q}} + a^{\textbf{q}} - t^{\textbf{q}}), \\
&= C \cdot \frac{1}{|Q|} \cdot 2(\Tilde{t}^{\textbf{q}'} + a^{\textbf{q}'} - t^{\textbf{q}'}) a^{\textbf{q}'},\\
where\\ 
C &= \frac{1}{2 \cdot \sqrt{\frac{1}{|Q|} \sum^{Q}_\textbf{q} (\hat{t}^{\textbf{q}}-t^{\textbf{q}})^2}}.
\end{align*}
Both partial derivatives are 0 if $a^{\textbf{q}'} = 0$, and this causes the chain rule to set all gradients of the parameters in $MLP_{\epsilon}$ as 0 for an arbitrary $\textbf{q}'$, and the back-propagation is impossible. To avoid the dying ReLU problem \cite{dyingrelu}, this loss function penalizes the depth overestimation of the last epsilon value instead of placing the ReLU activation at the network's end. An alternative to the proposed loss function is $L_\epsilon = min(0, \epsilon^{\textbf{q}})^2$, which is differentiable for all $\epsilon^{\textbf{q}}$. However, Equation \ref{eq:L_epsilon} is chosen to penalize negative $\epsilon^{\textbf{q}}$ values equally to strictly follow the sphere tracing algorithm \cite{sphere_tracing}, where the ray cannot penetrate the surface. Sub-gradient is used when $\epsilon^\textbf{q} = 0$, where $\frac{\partial L_\epsilon^{\textbf{q}}}{\partial \epsilon^\textbf{q}}$ is not differentiable.

\section*{Results on the PU1K and PU-GAN datasets}

Qualitative results on synthetic point clouds of PU1K \cite{pu_gcn} and PU-GAN \cite{pu_gan} datasets used in previous computer vision studies \cite{pu_net, mpu, pu_gan, qian2020pugeo, pu_gcn, disentangled, meta-pu, sspu, spu_net, pugac, pu_transformer, neural-points, gradpu, sapcu} supplement the experimental results in the main paper.
 
\begin{figure*}[!htb]
\centering
\includegraphics[width=\textwidth]{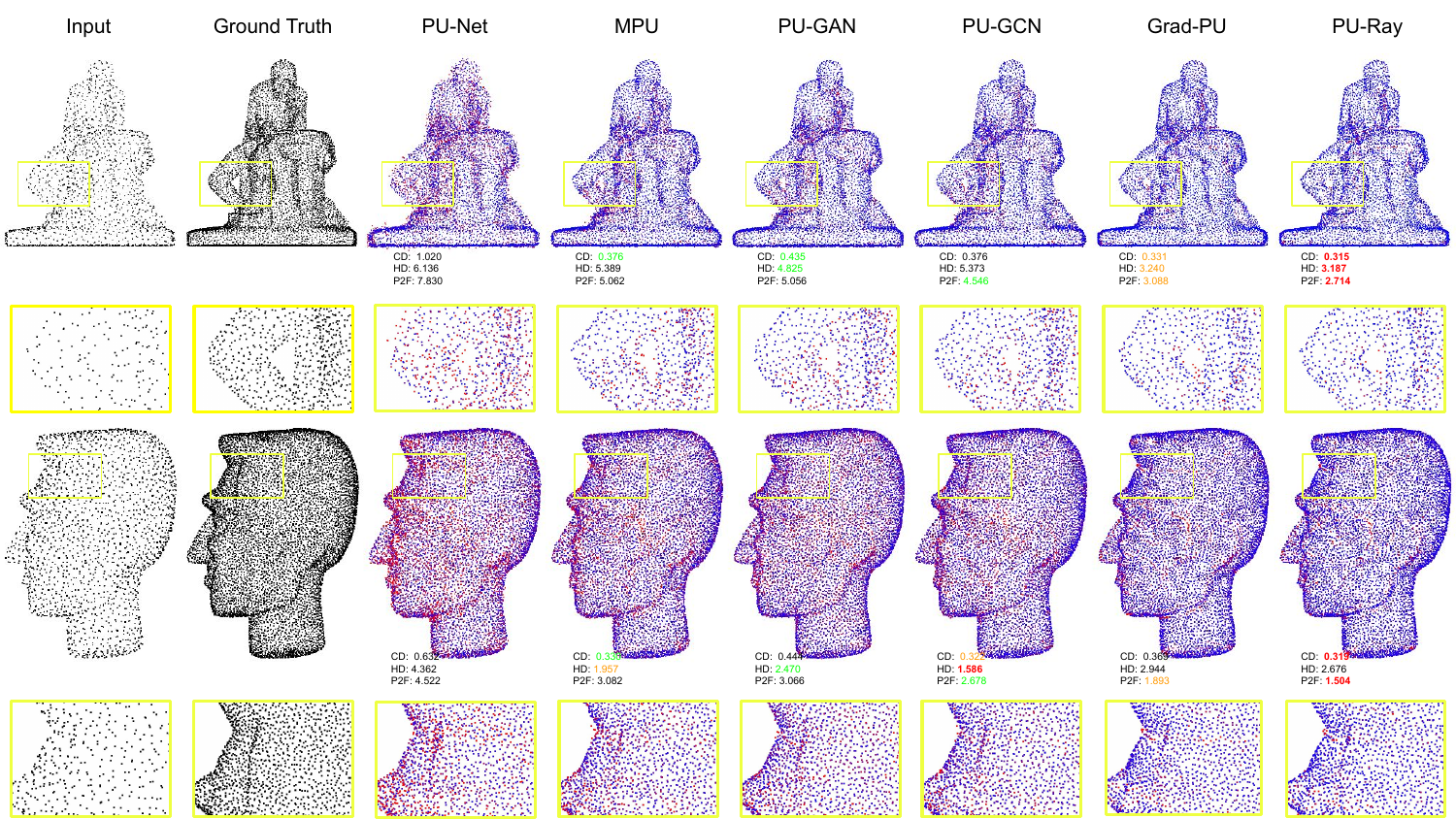}
\caption{Qualitative comparisons between state-of-the-art methods and PU-Ray for $4\times$ upsampling. Point colours are scaled from smaller (blue) to greater (red) P2F distances. The best three results are coloured with red (first), orange (second) and green (third).}
\label{fig:4x}
\end{figure*}

We observe that smaller P2F distances occur on complex surfaces. The current models fail to determine whether the space in between (e.g. triangle surrounded by a leg in the top row of Fig. \ref{fig:4x}) should be an implicit surface or a void. On the contrary, our method's query generation limits where the rays should be directed so that they will likely hit the surface. Our superior depth prediction gives smaller numbers of P2F errors on a folded surface, as shown along the hairline of a statue in the bottom row of Fig. \ref{fig:4x}.

\begin{figure}[!htb]
\centering
\includegraphics[width=\columnwidth]{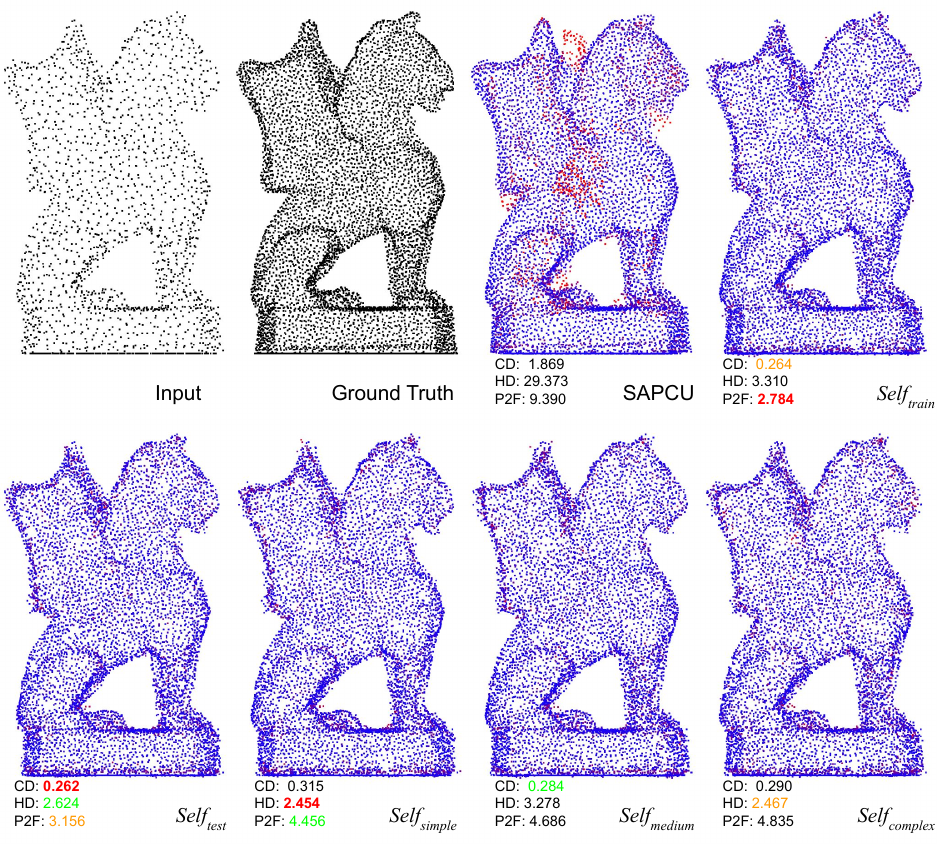}
\caption{Qualitative comparisons between SAPCU \cite{sapcu} and our self-supervised models for $4\times$ upsampling. Point colours are scaled from smaller (blue) to greater (red) P2F distances. The best three results are coloured with red (first), orange (second) and green (third).}
\label{fig:self_supervised}
\end{figure}

Fig. \ref{fig:self_supervised} illustrate how SAPCU fails with the PU-GAN test dataset, demonstrating the domain dependency issue. At the same time, all of our self-supervised methods can produce upsampling results with acceptable qualities.

\begin{figure}[!htb]
\centering
\includegraphics[width=\columnwidth]{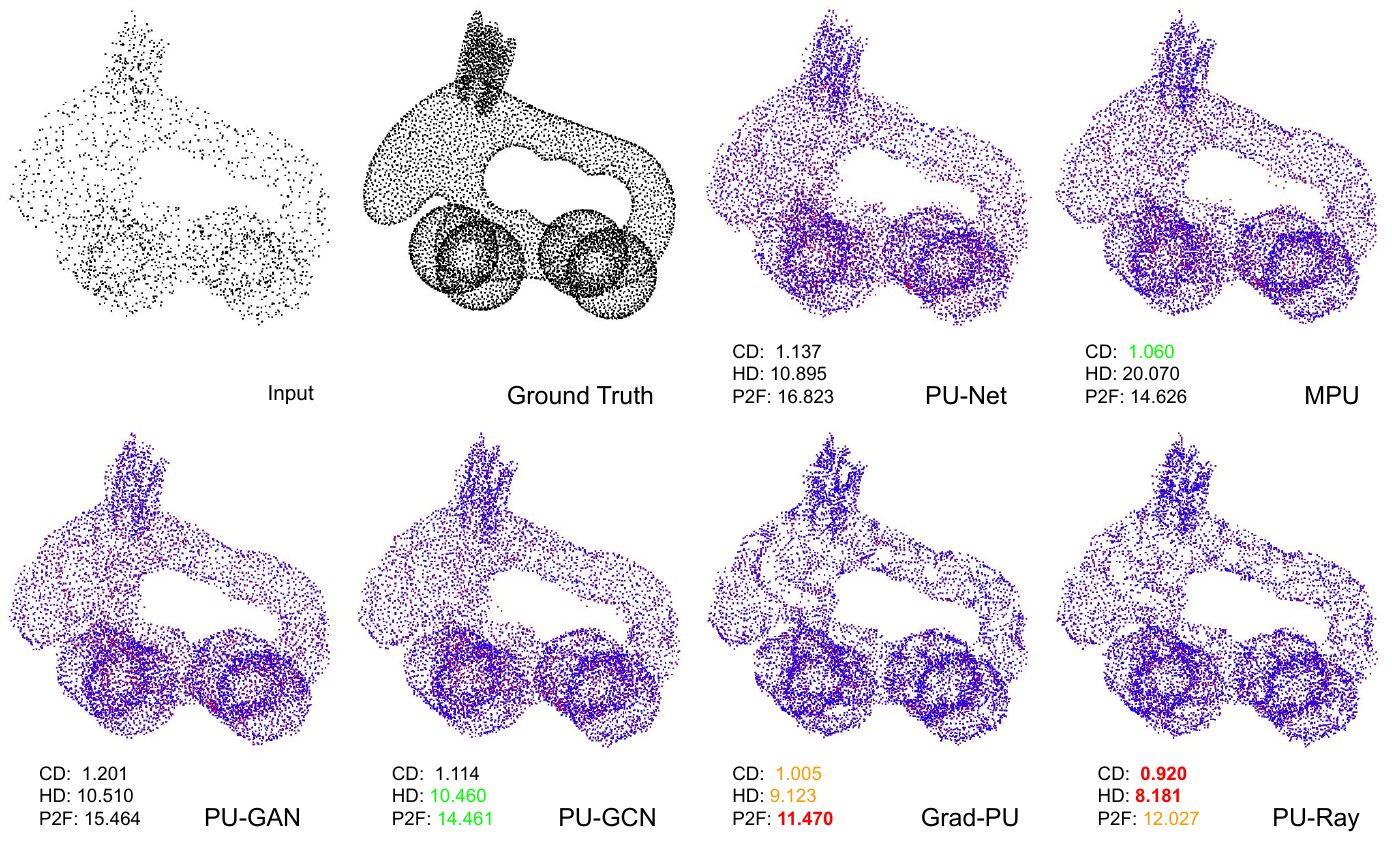}
\caption{Comparisons between upsampling results of existing methods and PU-Ray on a noisy point cloud input with $\gamma=0.02$ for $4\times$ upsampling. Our method is trained with supervised learning. The best three results are coloured with red (first), orange (second) and green (third).}
\label{fig:noise}
\end{figure}

\begin{figure}[!htb]
\centering
\includegraphics[width=\columnwidth]{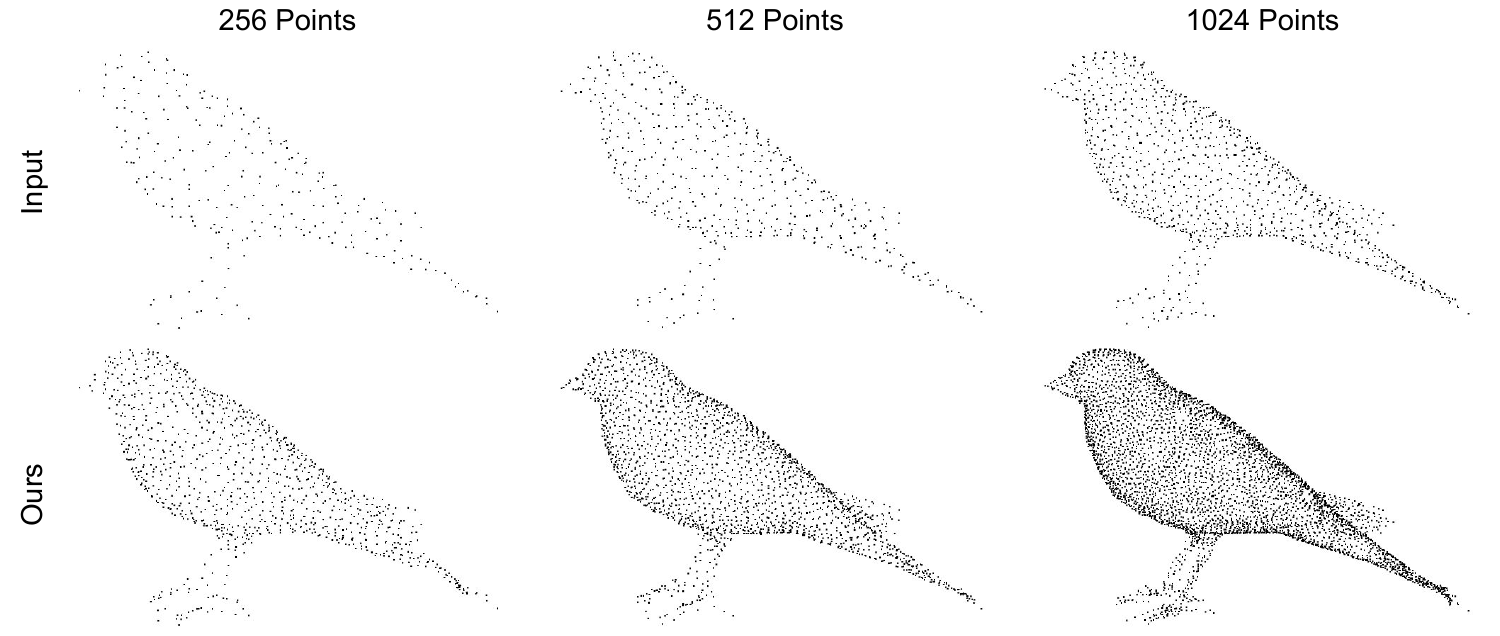}
\caption{PU-Ray's 4$\times$ upsampling outputs given different input sizes of 256, 512 and 1024.}
\label{fig:diff_input_size}
\end{figure}

Fig. \ref{fig:diff_input_size} illustrates that the shape is preserved regardless of the input sizes ranging from 4 to 2 times smaller than the training point clouds.

\section*{Qualitative Results on the KITTI-360 Dataset}
We present more of our qualitative results on snippets of the KITTI-360 dataset \cite{kitti360}. Points are generated in the unknown regions, illustrating the importance of user-controlled upsampling.

\begin{figure*}[!htb]
\centering
\includegraphics[width=\textwidth]{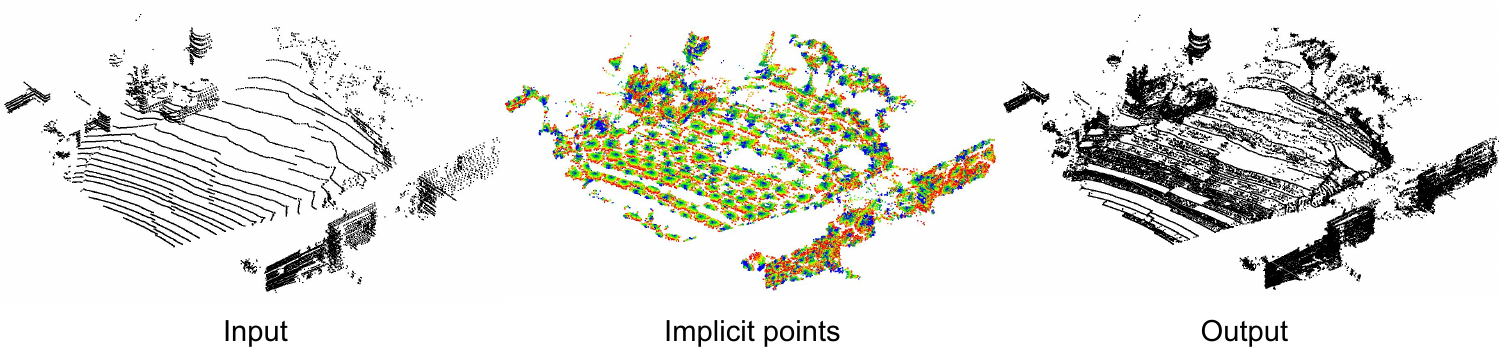}
\caption{Qualitative result on a KITTI-360 snippet.}
\label{fig:0000}
\end{figure*}
\begin{figure*}[!htb]
\centering
\includegraphics[width=\textwidth]{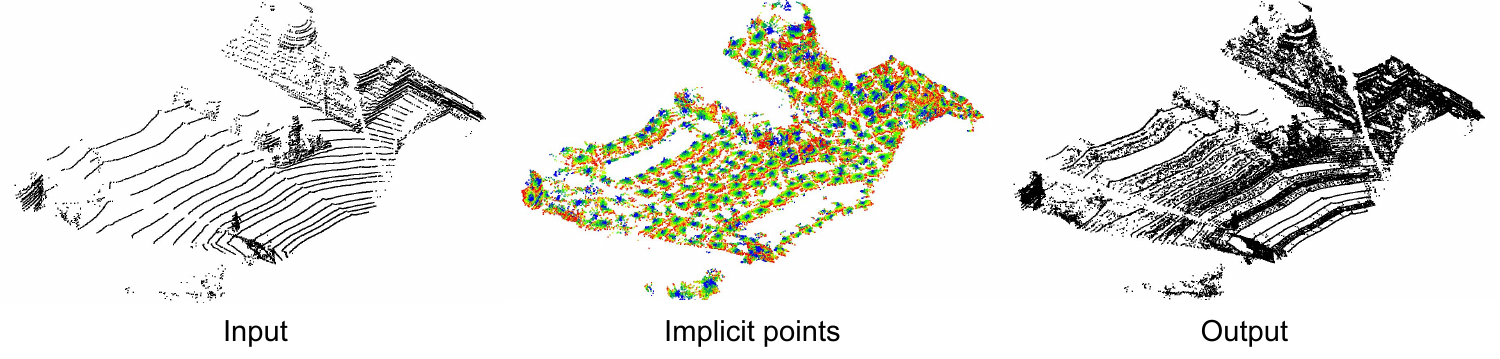}
\caption{Qualitative result on a KITTI-360 snippet.}
\label{fig:0002}
\end{figure*}
\begin{figure*}[!htb]
\centering
\includegraphics[width=\textwidth]{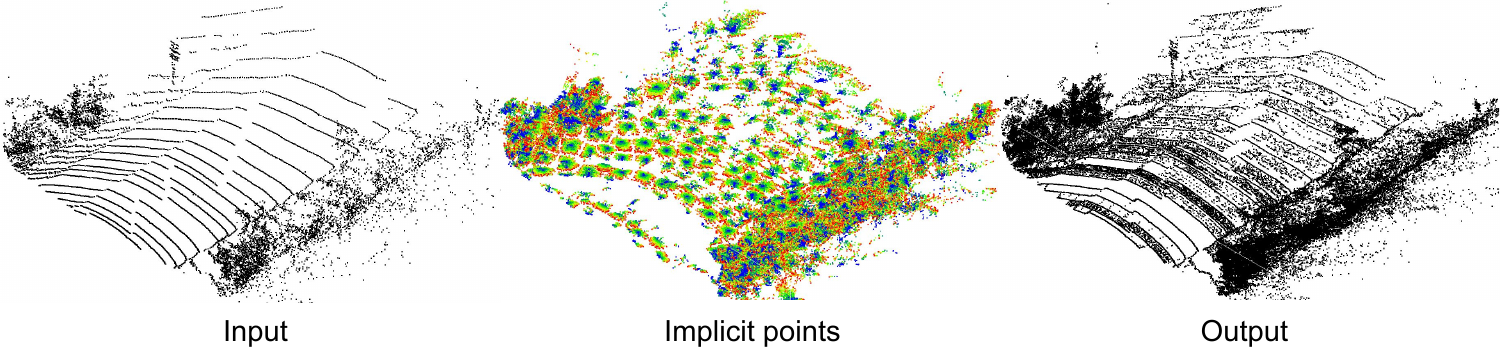}
\caption{Qualitative result on a KITTI-360 snippet.}
\label{fig:0003}
\end{figure*}
\begin{figure*}[!htb]
\centering
\includegraphics[width=\textwidth]{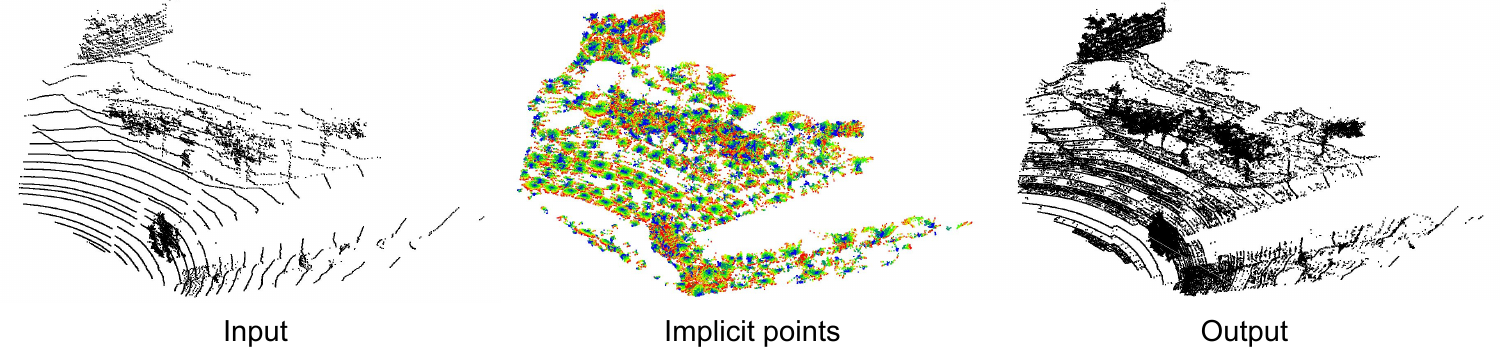}
\caption{Qualitative result on a KITTI-360 snippet.}
\label{fig:0004}
\end{figure*}
\begin{figure*}[!htb]
\centering
\includegraphics[width=\textwidth]{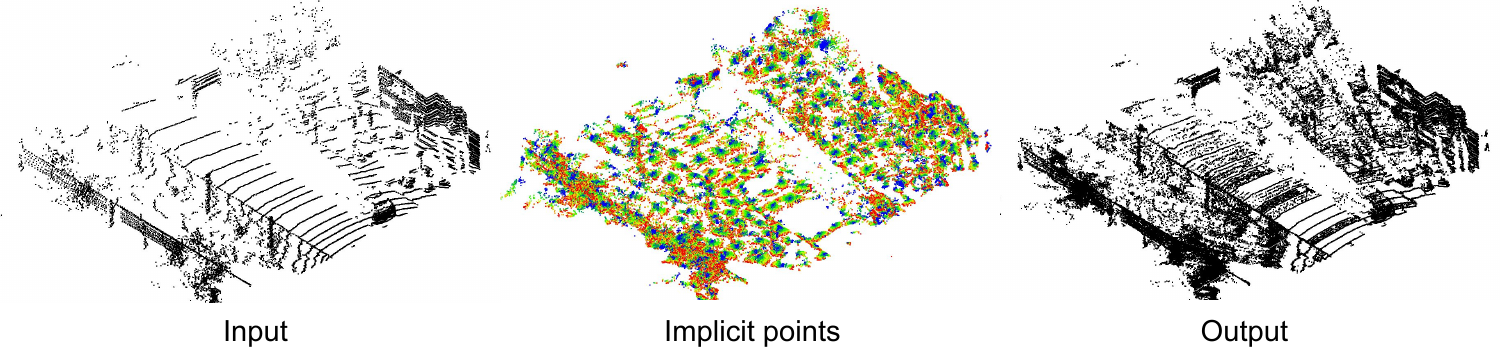}
\caption{Qualitative result on a KITTI-360 snippet.}
\label{fig:0005}
\end{figure*}
\begin{figure*}[!htb]
\centering
\includegraphics[width=\textwidth]{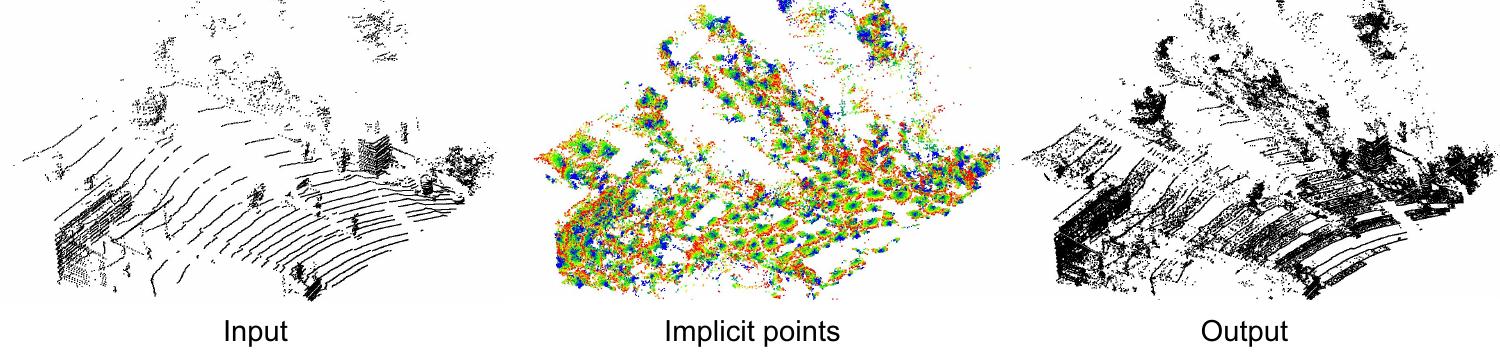}
\caption{Qualitative result on a KITTI-360 snippet.}
\label{fig:0006}
\end{figure*}
\begin{figure*}[!htb]
\centering
\includegraphics[width=\textwidth]{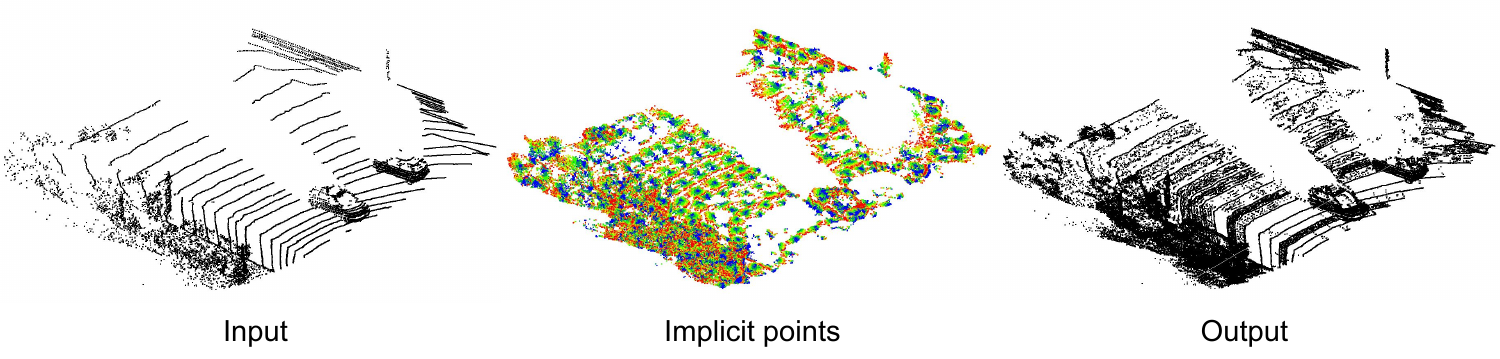}
\caption{Qualitative result on a KITTI-360 snippet.}
\label{fig:0007}
\end{figure*}
\begin{figure*}[!htb]
\centering
\includegraphics[width=\textwidth]{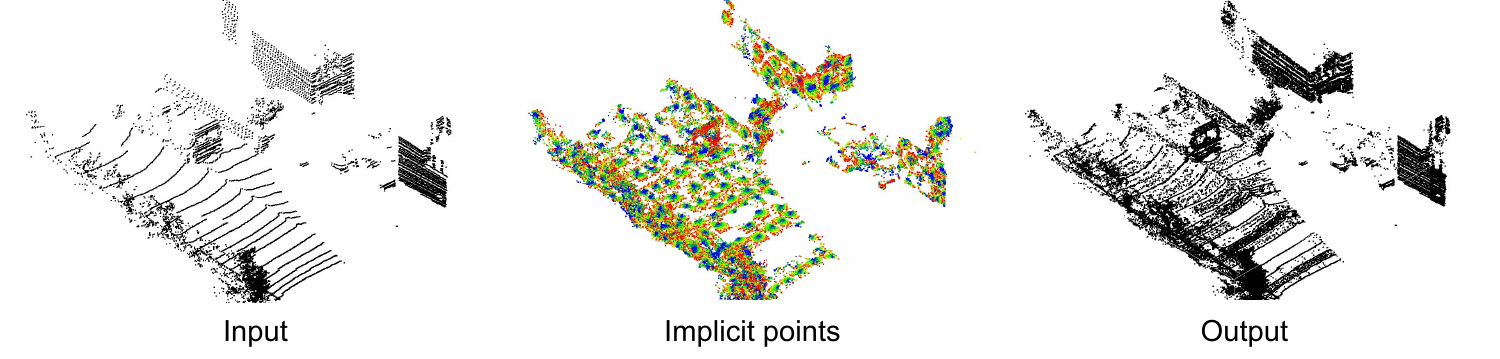}
\caption{Qualitative result on a KITTI-360 snippet.}
\label{fig:0009}
\end{figure*}
\begin{figure*}[!htb]
\centering
\includegraphics[width=\textwidth]{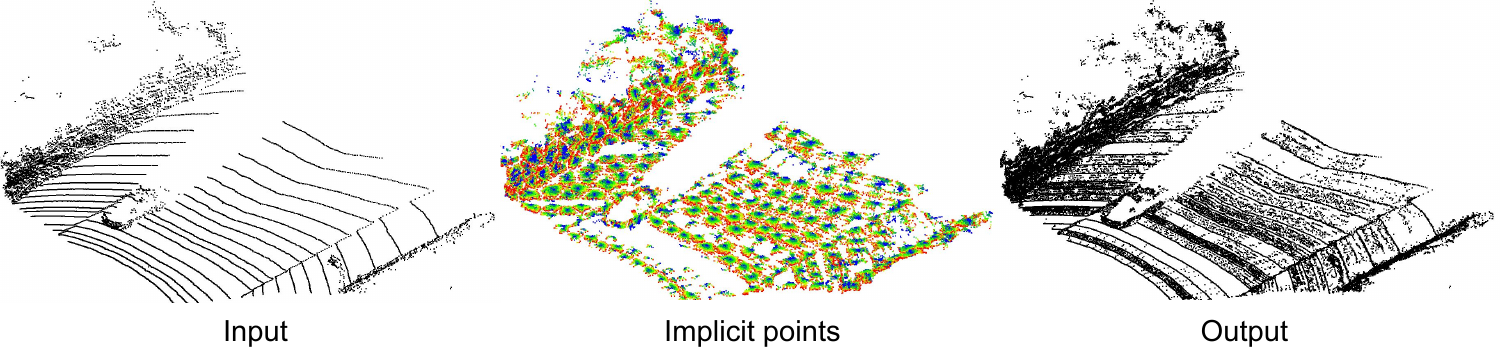}
\caption{Qualitative result on a KITTI-360 snippet.}
\label{fig:fig:0010}
\end{figure*}
}

\bibliographystyle{IEEEtran}
\bibliography{puray}